\documentclass[12pt]{article}
\pdfoutput=1

\usepackage[margin=1in]{geometry}
\usepackage{amsmath}
\usepackage{amssymb}
\usepackage[T1]{fontenc}
\usepackage{graphicx}
\usepackage{booktabs}
\usepackage{xltabular}
\usepackage[numbers,sort&compress]{natbib}
\usepackage{authblk}

\title{\textbf{Sequential Contextual Fit Predicts Human Behavioural and Neural Dynamics Across Domains}}
\author[1]{Kun Sun}
\author[2]{Rong Wang}
\affil[1]{Department of Linguistics, Tongji University \thanks{kunsun@tongji.edu.cn}}
\affil[2]{Department of Computational Linguistics, Tuebingen University}

\begin{document}
\maketitle

\begin{abstract}
Human perception, action and decision making unfold in sequences, but computational predictors are often domain-specific. This study computes and tests sequential contextual fit (SCF), an embedding-based measure of how well a current information state matches its recent context. The metric uses a simple recency-weighted similarity kernel and can be applied to words, sounds, visual scenes, affective states, choices, actions and neural representations. Across language processing, music-evoked emotion, a subset of audiovisual emotion EEG data, gambling decisions, human activity recognition and decision-related EEG, SCF predicted longer processing times, larger affective or behavioural transitions and stronger neural-state changes, with the direction and shape of the association varying by domain. These effects remained after controlling for established predictors including surprisal, reinforcement-learning prediction error, acoustic change, visual change and sensor change. SCF therefore provides a computational measurement layer for relating contextual compatibility to behavioural processing and cognitive/neural state-transition dynamics.
\end{abstract}

% Nature Computational Science Article structure:
% Introduction has no heading. Results and Online Methods may have topical subheadings.

\section{Introduction}

Human perception, action and decision making unfold in ordered streams. During reading, listening, music, social interaction, choice and movement, a current event is processed relative to recently active states. This local dependence shapes integration, expectation and updating, but computational analyses often rely on domain-specific quantities. A central question for computational science is whether one interpretable operation can measure contextual compatibility across sequential systems without assuming that all domains share the same feature space.
%Human behaviour and cognition often unfold in sequences. Words arrive during reading and listening, affective responses evolve during music and film, choices depend on recent outcomes, and actions are organized into temporally extended routines. In each case, the current input is interpreted against a recent context that shapes expectation, integration and state updating. A central challenge for computational science is to identify quantities that capture this local context dependence while remaining applicable beyond a single task or measurement modality.

Several frameworks capture important parts of this problem. In language, surprisal measures the negative log probability of a word given its preceding context \cite{hale2001probabilistic,levy2008expectation}. In decision making, reinforcement-learning prediction error measures the discrepancy between expected and obtained outcomes \cite{schultz1997neural,garrison2013prediction}. Predictive-processing and event-segmentation accounts likewise emphasize mismatch between incoming evidence and an active model of the current situation \cite{friston2010free,clark2013whatever,zacks2007event,zacks2007eventperception,baldassano2017eventstructure}. These approaches are powerful, but they answer different questions and require different inputs: surprisal requires a fitted probability model, prediction error requires values and outcomes, event-segmentation methods identify boundaries or latent event structure, and hidden-state or Markov models estimate an explicit transition system \cite{rabiner1989tutorial}. A physical or representational change score can quantify adjacent-unit discontinuity, but it does not by itself encode how the current unit relates to several earlier units \cite{foote2000automatic}. Thus, existing measures are not a common computational currency for comparing contextual compatibility across domains. Representational relatedness provides a complementary level of analysis. Once sequential units are mapped into vectors, their relations can be measured directly \cite{landauerdumais1997solution,kriegeskorte2008representational}.

The present study introduces \textbf{sequential contextual fit} (SCF), a recency-weighted embedding-space measurement operator for current--context compatibility. \textbf{SCF} computes the relatedness between the current unit and each of the preceding $K$ units, giving greater weight to more recent context. High SCF indicates representational continuity, whereas low SCF indicates contextual mismatch and may signal greater pressure for state updating.%High SCF indicates representational continuity, and low SCF indicates contextual mismatch and greater pressure for state updating. 
It transforms domain-specific sequence representations into transparent, unit-level metrics of contextual compatibility that can be evaluated against cognitive, behavioural and neural responses. %Instead, it transforms domain-specific sequence representations into transparent, unit-level predictors of cognitive, behavioural and neural processing across language, emotion, decision-making, action and other sequential domains. 
Its calculation does not require a domain-specific probability, reward or latent-state model, although the embedding used as input may itself be learned. Its generality therefore lies in the operator, not in a shared feature space or a claim that all domains have identical mechanisms. The intended mechanism is deliberately modest. Compatibility with recent context may support continuity, whereas low compatibility may increase the need to update the active representation. This is a measurement hypothesis rather than a claim that all domains implement a shared latent buffer, and it connects SCF to temporal-context and event-segmentation accounts \citep{howardkahana2002distributed,kurbyzacks2008segmentation}. The raw SCF score can be used directly as an interpretable continuity or mismatch signal. Statistical tests are required when the aim is to validate its empirical relationship with noisy human responses, uncertainty, nonlinearities and competing predictors.

We test whether lower SCF predicts greater processing cost, larger behavioural or affective transitions and stronger neural-state updating, and whether it contributes beyond established domain baselines including surprisal, reinforcement-learning prediction error, acoustic change, visual change and sensor change. We evaluate these predictions in language reading, EEG and fMRI, music-evoked emotion, audiovisual emotion EEG, gambling decisions, human activity recognition and decision-related EEG. %The resulting analyses will provide convergent predictive evidence for a common contextual-fit measure while also testing where alternative embedding operators provide a better description of a domain.
The present study could make three linked contributions. First, it provides a unified computational operation for relating each current state to its recent context while preserving domain-specific sequence units and representations. Second, it tests whether SCF explains variation beyond strong domain-specific baselines across diverse cognitive and neural activities, rather than treating embedding-based relatedness alone as sufficient evidence. Third, it evaluates the boundary conditions of the operator by comparing recency-weighted, unweighted and centroid-based alternatives. The aim is therefore a reusable measurement framework for testing continuity and updating in sequential systems, not a claim that one fitted model or one embedding geometry is optimal in every domain.

\section{Methods}

\subsection{Analysis overview}

All analyses used existing public datasets and treated behaviour, stimulus streams and neural recordings as ordered sequences of units. A unit was defined at the temporal resolution appropriate for each domain: a word in the language analyses, a 2 s audio segment in the music-emotion analysis, a 1 s visual/EEG segment in the audiovisual emotion-EEG analysis, a single trial in the gambling analyses, a 2.56 s smartphone-sensor window in the action analysis and a 0--2 s single-trial EEG segment in the decision-EEG analysis. For each unit, we constructed a domain-specific vector representation and investigated how well the current vector matched the recent context. Analyses were performed at the observation level after excluding rows with missing response values, missing predictors or insufficient preceding context for the specified window size. Continuous predictors were standardized within each analysis dataset before model fitting. Categorical variables were entered as factors. Unless otherwise stated, statistical models used generalized additive mixed models (GAMMs), so that non-linear predictor effects could be estimated without imposing a linear response shape. To demonstrate the robustness and generalizability of the method, we evaluated it using diverse representation types, including static pretrained vectors, contextualized embeddings generated by language models, and autoencoder-derived embeddings.

Table~\ref{tab:dataset-method-summary} summarizes the datasets, sequence units, embedding models, main baselines, controls and statistical models used across domains.

{\scriptsize
	\setlength{\tabcolsep}{2pt}
	\renewcommand{\arraystretch}{1.05}
	
	\begin{xltabular}{\textwidth}{
			>{\raggedright\arraybackslash}p{0.13\textwidth}
			>{\raggedright\arraybackslash}p{0.17\textwidth}
			>{\raggedright\arraybackslash}p{0.14\textwidth}
			>{\raggedright\arraybackslash}X
			>{\raggedright\arraybackslash}p{0.17\textwidth}
			>{\raggedright\arraybackslash}p{0.17\textwidth}
		}
		
		\caption{\textbf{Summary of datasets, representations and statistical models.}
			Each analysis converted an ordered stream into unit embeddings, computed SCF from recent context and tested its contribution beyond a domain-specific baseline and standard controls.}
		\label{tab:dataset-method-summary} \\
		
		\toprule
		\textbf{Domain} &
		\textbf{Dataset/source} &
		\textbf{Unit} &
		\textbf{Embedding or representation} &
		\textbf{Baseline and controls} &
		\textbf{Statistical model} \\
		\specialrule{0.35pt}{1pt}{1pt}
		\endfirsthead
		
		\multicolumn{6}{c}{\tablename\ \thetable{} continued from previous page} \\
		\toprule
		\textbf{Domain} &
		\textbf{Dataset/source} &
		\textbf{Unit} &
		\textbf{Embedding or representation} &
		\textbf{Baseline and controls} &
		\textbf{Statistical model} \\
		\specialrule{0.35pt}{1pt}{1pt}
		\endhead
		
		\specialrule{0.35pt}{1pt}{1pt}
		\multicolumn{6}{r}{Continued on next page} \\
		\endfoot
		
		\bottomrule
		\endlastfoot
		
		Language, eye movements &
		MECO \citep{siegelman2022meco} &
		Word fixation observation &
		Static lexical-semantic word embeddings \citep{Bojanowski2017fasttext} &
		Transformer surprisal \citep{radford2019language,goodkind-bicknell-2018-predictive,wilcox2020neural}, word frequency and word length &
		GAMMs \citep{wood2017generalized} for first fixation, gaze and total fixation durations \\
		\specialrule{0.35pt}{1pt}{1pt}
		
		Language, EEG &
		DERCo \citep{quach2024derco} &
		Word-locked EEG epoch &
		Word embeddings \citep{Bojanowski2017fasttext} &
		GPT-based surprisal \citep{radford2019language} and lexical controls &
		ERP regression and GAMMs \citep{wood2017generalized} for N400/P600 windows \citep{kutashillyard1980reading,osterhoutholcomb1992event}; FDR correction \citep{benjaminihochberg1995fdr} \\
		\specialrule{0.35pt}{1pt}{1pt}
		
		Language, fMRI &
		Alice \citep{bhattasali-etal-2020-alice} and Moth \citep{lebel2023natural} datasets &
		Word-aligned BOLD response &
		Word embeddings and transformer-based representations \citep{caucheteuxking2022brain,radford2019language} &
		GPT-2/GPT-Neo surprisal \citep{radford2019language} and timing controls &
		ROI-level GAMMs \citep{wood2017generalized} and FIR analyses; FDR correction \citep{benjaminihochberg1995fdr} \\
		\specialrule{0.35pt}{1pt}{1pt}
		
		Emotion &
		DEAM \citep{Aljanaki2017deam} &
		Two-second music unit at rater level &
		MERT audio embedding \citep{li2024mert} &
		Acoustic change \citep{foote2000automatic}, current valence and current arousal &
		Gaussian GAMM \citep{wood2017generalized}; song and rater random effects \\
		\specialrule{0.35pt}{1pt}{1pt}
		
		Emotion EEG &
		EAV \citep{lee2024eav} &
		One-second video/EEG segment &
		CLIP visual embedding \citep{radford2021clip} and EEG transition features &
		Visual embedding change, previous EEG transition, time, task and emotion &
		Gaussian GAMM \citep{wood2017generalized}; subject and interaction random effects; channel-wise FDR correction \citep{benjaminihochberg1995fdr} \\
		\specialrule{0.35pt}{1pt}{1pt}
		
		Decision making &
		Many Labs Iowa Gambling Task \citep{steingroever2015manylabs} &
		Trial &
		RL-derived latent decision-state vector \citep{rescorlawagner1972theory,suttonbarto1998reinforcement} &
		Negative RL prediction error \citep{rescorlawagner1972theory,suttonbarto1998reinforcement}, choice, outcome and trial controls &
		Binomial GAMM \citep{wood2017generalized} for next-trial switching; participant and study random effects \\
		\specialrule{0.35pt}{1pt}{1pt}
		
		Action &
		UCI HAPT \citep{reyesortiz2016transition,anguita2013public} &
		2.56-second sensor window &
		Autoencoder-32 embedding from 561 standardized features \citep{varamin2018deep} &
		L2 sensor change, previous activity and temporal controls &
		Binomial GAMM \citep{wood2017generalized} for transition status; subject random effect \\
		\specialrule{0.35pt}{1pt}{1pt}
		
		Decision EEG &
		Public IGT EEG dataset \citep{chavez2026iowa} &
		Single-trial 0--2-second EEG segment &
		PCA-20 neural embedding \citep{jolliffecadima2016pca,parra2005recipes} and RL-latent SCF representation \citep{rescorlawagner1972theory,suttonbarto1998reinforcement} &
		Negative RL prediction error, outcome, loss, trial and choice controls &
		Gaussian GAMM \citep{wood2017generalized} for EEG transition; channel-wise and time $\times$ channel FDR analyses \citep{benjaminihochberg1995fdr} \\
		
\end{xltabular}}

\subsection{Sequential contextual fit and one-step diagnostic}

SCF draws on several established traditions in context-sensitive and similarity-based modelling. Temporal-context models use a recency-sensitive, drifting context representation to explain memory dynamics \citep{howardkahana2002distributed}. SCF is conceptually related to these models because both emphasize recent contextual information, but SCF directly computes current-to-context similarity in an embedding space rather than estimating a drifting latent memory state. Latent semantic analysis established cosine similarity in a learned vector space as a measure of semantic relatedness \citep{landauerdumais1997solution}; spreading-activation theory linked processing ease to activation from semantic neighbours \citep{collinsloftus1975spreading}; representational similarity analysis compared structures across representational spaces \citep{kriegeskorte2008representational}; and audio novelty detection used local self-similarity to identify structural changes \citep{foote2000automatic}. SCF also differs from hidden Markov and related Markov-transition models, which estimate latent states and transition or emission probabilities from sequential data \citep{rabiner1989tutorial}. SCF is a deterministic measurement of how well the current observed representation matches a finite recent context. These frameworks are not restricted to single domains, but they motivate distinct computational operations. To our knowledge, no previous study has applied this specific recency-weighted embedding-space operator as a common measurement layer across these behavioural and neural domains. The following formalizes SCF. More differences are detailed in the Supplementary Material.

Let $z_t$ denote the vector representation of the current unit and $z_{t-i}$ the representation of the $i$th preceding unit from the same sequence. The primary SCF metric was the recency-weighted average of cosine similarities between the current unit and the previous $K$ units:
\begin{equation}
	\mathrm{SCF}_t = \frac{\sum_{i=1}^{K} a_i\,\mathrm{cos}(z_t,z_{t-i})}{\sum_{i=1}^{K} a_i},
\end{equation}
where
\begin{equation}
	a_i=\frac{K-i+1}{K}.
\end{equation}
The immediately preceding unit therefore receives the largest weight, and progressively older units receive smaller weights. The linear kernel was chosen as the primary specification because it preserves order without estimating a decay parameter; uniform and exponential alternatives were treated as sensitivity specifications rather than selected from the response data. Cosine similarity was computed as
\begin{equation}
	\mathrm{cos}(z_u,z_v)=\frac{z_u^\top z_v}{\lVert z_u\rVert_2\lVert z_v\rVert_2}.
\end{equation}
Higher SCF values indicate stronger local compatibility between the current unit and the recent context. Lower values indicate contextual mismatch or transition pressure. Some implementation scripts stored the unnormalised weighted sum, $\sum_i a_i\mathrm{cos}(z_t,z_{t-i})$. For a fixed window size this differs from the weighted average only by a constant factor, and all SCF predictors were standardized before modelling; the two parameterizations therefore give the same statistical ordering.

For the normalized formulation, $\mathrm{SCF}_t$ lies in $[-1,1]$ because it is a positive weighted average of cosine similarities. The cosine term is invariant to positive rescaling of either vector, and the metric requires no fitted parameters beyond the selected window and kernel. For embedding dimension $d$, computing one SCF value requires $O(Kd)$ arithmetic operations. This refers to the SCF calculation itself; end-to-end cost can still be dominated by signal preprocessing or embedding extraction. These properties make SCF interpretable as a deterministic measurement layer rather than a learned representation model. Because cosine similarity is conditional on the geometry and preprocessing of each embedding space, no single cross-domain isotropy transformation was imposed, and reported effects are therefore conditional on the domain-specific representation.

The primary window size was chosen before each main analysis according to the natural temporal scale of the domain. Language analyses used four-word windows; DEAM emotion used four preceding 2 s audio units; EAV audiovisual EEG used four preceding 1 s visual units; the decision, action and decision-EEG analyses used five preceding trials or sensor windows. Robustness analyses refitted the emotion, decision and action models with $K=3$, $K=4$ and $K=5$.

As a one-step diagnostic, we also computed
\begin{equation}
	\mathrm{SCF}^{(1)}_t=\cos(z_t,z_{t-1}),
\end{equation}
which is algebraically identical to the normalized SCF. This comparison distinguishes immediate embedding similarity from information accumulated across several preceding units. It was evaluated with the same controls, random-effect structure and grouped cross-validation used for the corresponding multi-unit analyses. The implementation details are seen in the Supplementary Material.

\subsection{Alternative SCF metrics}

To test whether effects were specific to the SCF operator rather than to the use of embeddings alone, we computed several alternative SCF metrics from the same vector spaces. First, an unweighted direct metric was defined as
\begin{equation}
	\mathrm{SCF}^{\mathrm{nowg}}_t=\sum_{i=1}^{K}\mathrm{cos}(z_t,z_{t-i}).
\end{equation}
Second, a pairwise-context variant added a weak measure of coherence within the preceding context:
\begin{equation}
	\mathrm{SCF}^{\mathrm{pair}}_t=\mathrm{SCF}_t+\gamma\sum_{i<j}b_{ij}\,\mathrm{cos}(z_{t-i},z_{t-j}),
\end{equation}
where $\gamma=0.15$ and the pairwise weights were smaller than the direct target-context weights. For four-word language windows, adjacent context pairs received weight 0.20, gap-2 pairs received weight 0.15 and gap-3 pairs received weight 0.10 before multiplication by $\gamma$. Third, centroid metrics compressed the preceding context into a single average vector \citep{snell2017networks},
\begin{equation}
	c_t=\frac{1}{K}\sum_{i=1}^{K}z_{t-i},
\end{equation}
and then measured either $\mathrm{cos}(z_t,c_t)$ or $\lVert z_t-c_t\rVert_2$. Centroid metrics quantify the proximity of the current unit to an aggregated representation of the preceding context, thereby collapsing the identity, temporal order and lag of individual context units. They are therefore more closely related to global-matching measures than to similarity-based interference, which specifically concerns competition among overlapping memory representations \citep{lewis2005activation,gordon2006similarity,vandyke2006retrieval}. By contrast, SCF retains lag-specific current-to-context relations through recency weighting. The two metric families may therefore capture different aspects of sequential structure and show distinct associations with behavioural and neural responses. The effects of these alternative metrics are reported in the Supplementary Material. 

\subsection{Language analyses}

Language was used as the largest testbed because strong baseline predictors are already available. The eye-movement analysis used the Multilingual Eye-Movement Corpus (MECO), which contains natural reading data from 13 languages \citep{siegelman2022meco}. The unit was a word-level fixation observation. Three response variables were analysed: first fixation duration, gaze duration and total fixation duration \citep{rayner1998eyemovements}. Word embeddings were static lexical-semantic vectors derived from the static pretrained \textit{fastText} \citep{Bojanowski2017fasttext}, rather than contextual embeddings from language models, to reduce the possibility that the method's effectiveness could be attributed to language-model-specific representations. SCF was computed as a local weighted similarity metric over the target word and nearby context words. Language implementations were dataset-specific. The legacy reading implementation used three preceding words and, for the reading analysis, one following word to approximate parafoveal preview; weights gave the immediately preceding word the largest contribution and distant context words smaller contributions, with pairwise similarities among preceding words forming a lower-weight coherence component. The present OneStop past-only diagnostic used the previous four words only and no following-word preview. Non-language analyses used preceding units only and did not include a following-unit term. %Surprisal was computed with transformer language models, using GPT- and BERT-based estimates in matched model sets. GAMMs included semantic coherence, surprisal, word length and word frequency, with repeated-observation structure represented by participant- and item/sentence-level random effects where available.

Surprisal, as the baseline metric, was computed with transformer language models, using GPT-based estimates in matched model sets. Surprisal quantifies the processing cost of a word as the negative log-probability of that word given its preceding context \citep{hale2001probabilistic,levy2008expectation}. Formally, surprisal for a target word $w_i$ is defined as:

\begin{equation}
	S(w_i) = -\log_2 P(w_i \mid w_1, w_2, \dots, w_{i-1})
	\label{eq:surprisal}
\end{equation}

\noindent where $P(w_i \mid w_1, \dots, w_{i-1})$ is the conditional 
probability assigned to word $w_i$ given the preceding words in the sentence. This quantity has been shown to correlate reliably with reading times and other online processing measures across a range of 
computational architectures, from $n$-gram and PCFG-based models \citep{hale2001probabilistic,levy2008expectation,demberg2008eyetracking} to connectionist and neural language models \citep{frank2011structure,smith2013surprisal}. More recent work has extended this line of research to transformer-based language models, including GPT-style unidirectional architectures and BERT-style masked or bidirectional architectures \citep{goodkind-bicknell-2018-predictive,wilcox2020neural}, which allow surprisal to be estimated with greater contextual sensitivity than earlier $n$-gram or recurrent approaches, such as GPT-2 \citep{radford2019language}.

%GAMMs included SCF, surprisal, word length and word frequency, with repeated-observation structure represented by participant- and item/sentence-level random effects where available. The GAMMs consists of the main predictor of our interest (SCF), the baseline metric (surprisal), controls (word length, word frequency) and random variables. Such setups are also applied in other domains. 
GAMMs included the primary predictor of interest, SCF, the baseline predictor, surprisal, and lexical controls for word length and word frequency. Where available, participant- and item- or sentence-level random effects were included to account for repeated observations. The same general modeling framework was applied across the other domains.

The language EEG analysis used the Dublin EEG-based Reading Experiment Corpus (DERCo) \citep{quach2024derco}. Word-locked EEG responses from 22 participants and 32 channels were analysed in N400 and P600 windows, motivated respectively by classic N400 and P600 findings \citep{kutashillyard1980reading,osterhoutholcomb1992event}. These time-localized analyses follow the broader use of temporal generalization to characterize evolving neural representations \citep{kingdehaene2014temporal}. The semantic-fit predictor was SCF, computed as a local embedding-based target-context fit measure over recent discourse context. The baseline predictor was GPT-based word surprisal. Channel-wise regression-based ERP analyses and GAMMs tested semantic relevance and surprisal while controlling lexical variables, including word frequency and word length, and repeated observations by participant. Effects were summarized separately for the N400 and P600 windows. Multiple comparisons across channels were controlled with the Benjamini--Hochberg false-discovery-rate procedure \citep{benjaminihochberg1995fdr}. In brief, SCF and surprisal were computed for the EEG and fMRI datasets using procedures analogous to those applied to the eye-movement datasets.%In short, the computation of SCF and surprisal is similar to the ones in eyemovement datasets. 

The language fMRI analyses used two naturalistic speech-comprehension datasets. In the Alice dataset \citep{bhattasali-etal-2020-alice}, 26 participants listened to the first chapter of \textit{Alice's Adventures in Wonderland}; the stimulus contained 2,129 words, fMRI data were sampled with TR = 2 s and the run contained 372 time points. In the Moth dataset \citep{lebel2023natural}, 8 participants listened to 27 autobiographical podcast stories, also acquired with TR = 2 s. Word surprisal was computed primarily with GPT-2, with GPT-Neo used as a robustness check in the Moth analysis. Semantic relevance was computed from word embeddings using the recent-word semantic-fit algorithm. Representation-based modelling of naturalistic brain responses provides a related motivation for evaluating learned embeddings against neural data \citep{caucheteuxking2022brain,millet2022realistic}. Models included lexical and timing controls where available, including word frequency and word length. Two complementary fMRI analyses were used. First, transformed BOLD responses were modelled with ROI-level GAMMs. Second, original continuous BOLD time series were analysed with finite-impulse-response (FIR)/deconvolution models to test whether semantic relevance and surprisal showed hemodynamically plausible delayed effects \citep{glover1999deconvolution}. Alice analyses used 12 predefined ROIs; Moth analyses used 30 analyzable ROIs. The Benjamini--Hochberg FDR procedure was applied across ROI-level tests \citep{benjaminihochberg1995fdr}.

\subsection{Emotion analyses}

The emotion analysis used the DEAM music-emotion dataset with continuous valence and arousal annotations \citep{russell1980circumplex,Aljanaki2017deam}. Songs were segmented into non-overlapping 2 s units. Each unit was represented with MERT audio embeddings \citep{li2024mert}, and SCF was computed by comparing the current 2 s unit with the preceding four 2 s units from the same song. Analyses were performed at the rater level after merging song-unit embeddings with per-rater valence/arousal annotations. The final dataset contained 194,777 observations, 1,802 songs, 108 raters and 19,822 song units. The baseline metric was immediate acoustic change. Local self-similarity and novelty detection motivate this comparator \citep{foote2000automatic}, while continuous musical-feature modelling and mechanistic accounts explain why time-varying acoustic structure can covary with perceived emotion \citep{schubert2004modeling,juslinvastfjall2008emotional}.

The response variable was next affective change, defined as the Euclidean displacement in valence-arousal space from the current unit to the next unit:
\begin{equation}
	\Delta A_{t+1}=\sqrt{(\mathrm{valence}_{t+1}-\mathrm{valence}_{t})^2+(\mathrm{arousal}_{t+1}-\mathrm{arousal}_{t})^2}.
\end{equation}
The main stimulus-side baseline was local MERT acoustic change,
\begin{equation}
	\mathrm{acoustic\ change}_t
	=
	1-\mathrm{cos}(m_t,m_{t-1}),
\end{equation}
where $m_t$ is the MERT embedding of the current 2-s music unit.
This measure operationalizes immediate acoustic novelty as displacement
between adjacent units in the same pretrained audio-representation space.
It is motivated by previous work showing that temporal variation in
acoustic and psychoacoustic features contributes to the prediction of
time-varying musical emotion
\citep{coutinho2011musical,schmidt2011modeling}.
The baseline therefore tests whether SCF predicts affective change beyond
immediate acoustic change in the same representational space. The full Gaussian GAMM was
\begin{align}
	\Delta A_{t+1} \sim &\ s(\mathrm{SCF},k=5)+s(\mathrm{acoustic\ change},k=5) \\
	&+s(\mathrm{valence}_t,k=5)+s(\mathrm{arousal}_t,k=5) \\
	&+s(\mathrm{song},\mathrm{bs}=\mathrm{re})+s(\mathrm{rater},\mathrm{bs}=\mathrm{re}).
\end{align}
Reduced models removed either SCF or acoustic change from the full model, and unique contribution was quantified as $\Delta\mathrm{AIC}=\mathrm{AIC}_{\mathrm{reduced}}-\mathrm{AIC}_{\mathrm{full}}$.

The audiovisual emotion-EEG analysis used the EAV dataset \citep{lee2024eav}, a multimodal emotional-conversation dataset with synchronized EEG, audio and video recordings from 42 participants \cite{lee2024eav}. Each participant contributed 200 interactions covering neutral, anger, happiness, sadness and calmness conditions. The EAV EEG montage contains 30 channels (Fp1, Fp2, F7, F3, Fz, F4, F8, FC5, FC1, FC2, FC6, T7, C3, Cz, C4, T8, CP5, CP1, CP2, CP6, P7, P3, Pz, P4, P8, PO9, O1, Oz, O2 and PO10), sampled at 500 Hz.

Each 20-s interaction was divided into 1-s units. The visual representation of each unit was the CLIP image embedding of its midpoint video frame \citep{radford2021clip}. The use of CLIP is motivated by
evidence that its visual embedding space contains information relevant to image emotion and visual sentiment \citep{bondielli2021leveraging}. Embeddings were $L_2$-normalized before SCF computation, and SCF was calculated from the previous four visual units within the same interaction. Five
embedding-context metrics were derived from the same CLIP vectors: weighted direct SCF, unweighted direct SCF, direct SCF with pairwise context coherence, centroid cosine and centroid $L_2$. The stimulus-side baseline was one-step visual change,
\begin{equation}
	\mathrm{visual\ change}_t
	=
	1-\mathrm{cos}(v_t,v_{t-1}),
\end{equation}
where $v_t$ is the CLIP embedding of the current visual unit. This measure served as a representation-matched comparator of immediate visual displacement between consecutive units, rather than as a previously established metric of affective change.

The EAV EEG response was constructed independently of the visual SCF predictor. For each 1 s EEG unit, we extracted channel-wise means, standard deviations and root-mean-square amplitudes from the raw 30-channel signal \citep{mensen2017eeg}. These descriptive features were defined for the present analysis rather than adopted as a standard emotion-EEG baseline. The EEG transition response was the L2 distance between the current EEG feature vector and the recency-weighted centroid of the previous four EEG feature vectors from the same interaction:
\begin{equation}
	\mathrm{EEG\ transition}_t=\left\lVert e_t-\frac{\sum_{i=1}^{4}a_i e_{t-i}}{\sum_{i=1}^{4}a_i}\right\rVert_2.
\end{equation}
The full Gaussian GAMM for the main EAV analysis was
\begin{align}
	\mathrm{EEG\ transition}_t \sim &\ s(\mathrm{SCF},k=5)+s(\mathrm{visual\ change},k=5) \\
	&+s(\mathrm{previous\ EEG\ transition},k=5)+s(\mathrm{time},k=5) \\
	&+\mathrm{task}+\mathrm{emotion}+s(\mathrm{subject},\mathrm{bs}=\mathrm{re}) \\
	&+s(\mathrm{interaction},\mathrm{bs}=\mathrm{re}).
\end{align}
Reduced models removed the target SCF variant while retaining the visual-change baseline and all controls \citep{ittibaldi2009bayesian}. Channel-wise partial-effect and topographic figures used the same predictors but omitted interaction random intercepts so that 30 channel-wise visualization models could be fitted tractably; these models were treated as spatial summaries rather than the primary inferential tests.

\subsection{Decision-making analysis}
\subsubsection{Behavioural analysis}

The decision analysis used the Many Labs Iowa Gambling Task behavioural dataset assembled from 10 studies \citep{steingroever2015manylabs}. The unit was one gambling trial, and the final analysis contained 62,206 trials from 617 participants across 10 studies. The response variable was next-trial switching:
\begin{equation}
	\mathrm{switch}_{t+1}=\mathbb{1}[\mathrm{choice}_{t+1}\ne\mathrm{choice}_t].
\end{equation}

To avoid direct leakage of raw task variables into the representation, the primary decision representation was defined as an RL-derived pre-outcome latent state. This approach follows computational models of the Iowa Gambling Task in which deck-specific expectations are updated sequentially from prediction errors and mapped to choice probabilities through a softmax
rule \citep{busemeyer2002contribution,ahn2008comparison}. For each participant, deck values were initialized at zero and updated sequentially with a fixed learning rate:
\begin{align}
	\mathrm{PE}_t &= r_t-Q_t(\mathrm{choice}_t),\\
	Q_{t+1}(\mathrm{choice}_t) &= Q_t(\mathrm{choice}_t)+0.1\,\mathrm{PE}_t.
\end{align}
A separate expected-loss state was updated as
\begin{equation}
	L_{t+1}(\mathrm{choice}_t)=L_t(\mathrm{choice}_t)+0.1\,(\mathrm{loss}_t-L_t(\mathrm{choice}_t)).
\end{equation}
The embedding for trial $t$ was computed before observing the current trial outcome. It contained expected net values for decks A--D, expected loss values for decks A--D, the chosen deck value, the chosen expected loss, value gap, loss gap, the softmax probability of the chosen deck, softmax entropy and a perseveration-prior indicator \citep{ahn2008comparison,ahn2011modelbased}. SCF was computed over the previous five pre-outcome latent decision-state vectors.

The main decision baseline was negative reinforcement-learning prediction error \citep{rescorlawagner1972theory,suttonbarto1998reinforcement},
\begin{equation}
	\mathrm{negative\ PE}_t=\max(-\mathrm{PE}_t,0),
\end{equation}
which captures worse-than-expected outcomes that should encourage behavioural updating. The full binomial GAMM was
\begin{align}
	\mathrm{switch}_{t+1} \sim &\ s(\mathrm{SCF},k=5)+s(\mathrm{negative\ PE},k=5)+\mathrm{choice}_t \\
	&+s(\mathrm{net\ outcome}_t,k=5)+s(\mathrm{trial},k=5) \\
	&+s(\mathrm{participant},\mathrm{bs}=\mathrm{re})+s(\mathrm{study},\mathrm{bs}=\mathrm{re}).
\end{align}
Reduced models removed either SCF or negative prediction error from the full model.

\subsubsection{Decision-related EEG analysis}

The decision-EEG analysis used a separate public dataset in which IGT behaviour and EEG were acquired simultaneously \citep{chavez2026iowa}. Recent ERP work provides a broader account of decision stages and neural components studied with the IGT \citep{latibeaudiere2025decision}. The analysis tested whether behavioural/RL contextual fit predicted neural-state updating. The public dataset contained processed EEG files for 55 participants. Each participant contributed 200 IGT trials, yielding 11,000 trials before context-window omission and 10,725 trials in the main GAMMs. Each EEG unit was the single-trial 0--2 s segment after the task marker. Signals were baseline-corrected using the -200 to 0 ms pre-marker interval, block-averaged from 256 Hz to 32 Hz and flattened across time and channels, following a linear multivariate EEG representation strategy \citep{parra2005recipes}. Dimensionality was then reduced with principal component analysis \citep{jolliffecadima2016pca}; the first 20 principal components were retained as a trial-level neural embedding.

The behavioural representation and prediction-error baseline were motivated by reinforcement-learning and IGT research \citep{schultz1997neural,bechara1994insensitivity}. The neural response variable itself was computed independently of the behavioural/RL SCF predictor. For trial $t$, the current EEG embedding was compared with the recency-weighted centroid of the previous five EEG embeddings from the same participant:
\begin{equation}
	\mathrm{EEG\ transition}_t=\left\lVert e_t-\frac{\sum_{i=1}^{5}a_i e_{t-i}}{\sum_{i=1}^{5}a_i}\right\rVert_2.
\end{equation}
The main predictor was the RL-latent SCF computed from behavioural decision-state embeddings, not from EEG embeddings. This separation avoids circularity between response construction and predictor construction. The full Gaussian GAMM was
\begin{align}
	\mathrm{EEG\ transition}_t \sim &\ s(\mathrm{RL\mbox{-}latent\ SCF},k=5)+s(\mathrm{negative\ PE},k=5) \\
	&+s(\mathrm{net\ outcome}_t,k=5)+\mathrm{loss}_t+s(\mathrm{trial},k=5) \\
	&+\mathrm{choice}_t+s(\mathrm{participant},\mathrm{bs}=\mathrm{re}).
\end{align}
Additional response variants used EEG cosine mismatch and EEG cosine fit, but the L2 transition response was the primary neural-state updating measure.

For scalp visualization, we computed channel-wise trial-level transition responses without averaging across trials. For each electrode, the current 0--2 s waveform was compared with a recency-weighted centroid of the previous five trial waveforms at that electrode. Channel-wise regressions estimated the unique SCF effect while controlling negative prediction error, net outcome, loss indicator, trial number, choice and participant fixed effects. The topographic maps plot regression coefficients and $t$ statistics across scalp channels and are used only to visualize scalp distribution, not to infer neural sources. For the time $\times$ channel heatmap, each time point and channel was analysed separately. The response was the absolute deviation between the current baseline-corrected waveform value and the recency-weighted recent-context waveform value. Each time $\times$ channel response was standardized and predicted by the same SCF, baseline and control variables. FDR correction was applied over the mass-univariate time $\times$ channel tests. ROI time courses were computed by averaging channel-level coefficients or statistics within predefined channel groups.

\subsection{Action analysis}

The primary action embedding was generated by an autoencoder trained on standardized 561-dimensional sensor-feature vectors. These features summarize time- and frequency-domain properties of smartphone accelerometer and gyroscope signals \citep{anguita2013public}. Autoencoders have previously been
used in sensor-based human activity recognition to learn compact latent representations through reconstruction-based training and dimensionality reduction \citep{varamin2018deep}. The architecture used here was
\begin{equation}
	561 \rightarrow 128 \rightarrow 32 \rightarrow 128 \rightarrow 561,
\end{equation}
with the 32-dimensional bottleneck vector used as the action-state embedding. The saved reconstruction mean squared error was 0.12456. The architecture, bottleneck dimensionality and reconstruction error were specific to the present analysis. SCF was computed between the current bottleneck vector and the previous five bottleneck vectors from the same subject and split-specific
sequence.

The main baseline was one-step sensor change,
\begin{equation}
	\mathrm{sensor\ change}_t
	=
	\left\lVert x_t-x_{t-1}\right\rVert_2,
\end{equation}
where $x_t$ is the standardized 561-dimensional sensor-feature vector. This representation-matched baseline quantifies immediate displacement between consecutive sensor states. It is motivated by work treating abrupt changes in sensor time series as boundaries between activity states
\citep{aminikhanghahi2017survey,reyesortiz2016transition}, but the particular Euclidean-distance formulation was defined for the present analysis.

The full binomial GAMM was
\begin{align}
	\mathrm{transition}_t \sim &\ s(\mathrm{SCF},k=5)+s(\mathrm{sensor\ change},k=5)+\mathrm{previous\ activity} \\
	&+s(\mathrm{window\ index},k=5)+\mathrm{split}+s(\mathrm{subject},\mathrm{bs}=\mathrm{re}).
\end{align}
The current activity label was not included as a control because it directly defines the transition response and would introduce label leakage. Reduced models removed either SCF or sensor change from the full model. In metric-specificity analyses, centroid cosine and centroid L2 were computed from the same autoencoder embeddings and included either as single metrics with the same baseline and controls or jointly with SCF.

\subsection{Statistical modelling and model comparison}

%GAMMs were fitted in R using smooth terms with basis dimension $k=5$ for continuous predictors unless otherwise stated. Gaussian models were used for continuous responses, including fixation duration, affective change and EEG transition magnitude \citep{wood2017generalized}. Binomial models were used for binary responses, including next-trial switching and action-transition status. Random effects were represented as penalized random-effect smooths, $s(\mathrm{factor},\mathrm{bs}=\mathrm{re})$. For channel-wise and time-resolved EEG visualization analyses, ordinary least-squares models with participant fixed effects were used to make the mass-univariate computations tractable and transparent. These random-effect terms account for clustered observations but do not by themselves remove within-sequence residual autocorrelation; targeted AR(1) and block-based sequence-robustness analyses were performed for DEAM and HAPT and are reported in the Appendix, while residual-correlation modelling was not applied uniformly to every domain.
GAMMs were fitted in R using smooth terms with basis dimension $k=5$ for continuous predictors unless otherwise stated. The general model specification was:

\begin{equation}
	\begin{aligned}
		\text{response} \sim\;&
		s(\text{main predictor}, k=5) \\
		&+ s(\text{baseline metric}, k=5) \\
		&+ \sum_j s(\text{controls}_j, k=5) \\
		&+ \sum_m s(\text{random factor}_m,
		\mathrm{bs}=\texttt{"re"}).
	\end{aligned}
\end{equation}

Gaussian models were used for continuous responses, including fixation duration, affective change and EEG transition magnitude \citep{wood2017generalized}, whereas binomial models were used for binary responses, including next-trial switching and action-transition status. Random effects were represented as penalized random-effect smooths, $s(\mathrm{factor},\mathrm{bs}=\texttt{"re"})$. For channel-wise and time-resolved EEG visualization analyses, ordinary least-squares models with participant fixed effects were used to make the mass-univariate computations tractable and transparent. These random-effect terms account for clustered observations but do not by themselves remove within-sequence residual autocorrelation. Targeted AR(1) and block-based sequence-robustness analyses were therefore performed for DEAM and HAPT and are reported in the Appendix, although residual-correlation modelling was not applied uniformly across all domains.

The main inferential test in each domain was whether SCF remained significant when entered together with a strong domain baseline and standard controls. For language, the baseline was surprisal. For DEAM music emotion, the baseline was MERT acoustic change. For EAV audiovisual emotion EEG, the baseline was one-step CLIP visual embedding change, with previous EEG transition included as a neural temporal control. For decision making, the baseline was negative RL prediction error. For action, the baseline was direct sensor change. For decision EEG, the behavioural/RL baseline was negative RL prediction error. Model summaries report F statistics for Gaussian smooths and approximate chi-square statistics for binomial smooths. Multiple comparisons in channel-, ROI- and time-resolved analyses were controlled using the Benjamini--Hochberg false-discovery-rate procedure \citep{benjaminihochberg1995fdr}.

Unique contribution was quantified by reduced-model comparison. For a target predictor $x$, we fitted the full model and a reduced model with $x$ removed but all other predictors retained, then computed
\begin{equation}
	\Delta\mathrm{AIC}_x=\mathrm{AIC}_{\mathrm{reduced},x}-\mathrm{AIC}_{\mathrm{full}}.
\end{equation}
Positive values of $\Delta\mathrm{AIC}$ indicate that removing the predictor worsened model fit, and larger values indicate a larger unique contribution conditional on the other terms in the model. AIC is an estimator of expected out-of-sample prediction error and therefore of relative model quality within a matched model set. These differences are interpreted only within the same dataset, response, likelihood and predictor set; their absolute values are not compared as a common effect-size scale across domains. Because embedding-derived metrics can be highly correlated, joint models containing SCF, unweighted SCF, pairwise SCF and centroid metrics were interpreted as conservative collinearity-sensitive tests. Single-metric models with the same baseline and controls were used to compare each metric under less severe collinearity.

The inferential hierarchy was also kept separate from the exploratory metric comparison. The recency-weighted direct SCF was the primary operator because it preserves the lag-specific current-to-context relation and has a fixed, transparent kernel. Unweighted, pairwise and centroid metrics were secondary or exploratory alternatives used to identify representation-dependent boundary conditions; they were not used to replace the primary SCF test after inspecting the results. Accordingly, the paper does not apply one global Bonferroni correction across all domains, responses and mechanistically different analysis families. Instead, false-discovery-rate correction was applied within the pre-defined channel-, ROI- and time-resolved families, while domain-level smooth tests are reported with their model-comparison and held-out prediction evidence.

\subsection{Grouped leave-subject-out validation and window sensitivity}

Prediction validation was conducted for the emotion, decision and action domains. We used grouped five-fold leave-subject-out cross-validation: complete subjects, raters or participants were held out from training, so no individual appeared in both the training and test data within a fold. The action and decision tasks used held-out AUC because the responses were binary. The emotion task used held-out $R^2$ because the response was continuous. Prediction models compared a baseline/control model with a baseline/control model plus SCF. Random intercepts were omitted from the held-out prediction models because held-out individuals have no fitted random-effect levels. The embedding step was not a supervised response-prediction model: pretrained checkpoints, unsupervised autoencoders and unsupervised PCA used only stimulus, sensor or EEG representation data and never used the behavioural or neural response, outcome label or model residual. Thus, fitting an embedding on all available covariates does not create response-label leakage. The resulting validation should nevertheless be interpreted as validation of a fixed representation and SCF measurement layer, rather than as a strict end-to-end estimate in which every representation-learning parameter is re-fitted within each training fold.

Window-size sensitivity was tested by recomputing direct weighted SCF with $K=1$, $K=3$, $K=4$ and $K=5$ in emotion, decision and action. For each $K$, the corresponding full GAMM was refitted with the same baseline, controls and random-effect structure used in the main analysis. Grouped five-fold cross-validation compared a matched baseline/control model with the same model plus SCF. This design tests both whether SCF contains information beyond adjacent-unit similarity and whether its predictive contribution depends on a single arbitrary multi-unit window.

The same representation separation was used in the decision-EEG analysis. Behavioural/RL-latent embeddings were computed from task history and pre-outcome state variables, whereas the EEG transition response was computed from the EEG signal itself. No EEG response, response residual or next-trial outcome was used to fit the behavioural embedding or the SCF predictor. This separation prevents circular response construction, although the shared RL state family means that SCF and negative prediction error are not assumed to be statistically independent; their joint-model interpretation is therefore incremental rather than causal.

\begin{figure}[!h]
	\centering
	\includegraphics[width=\textwidth]{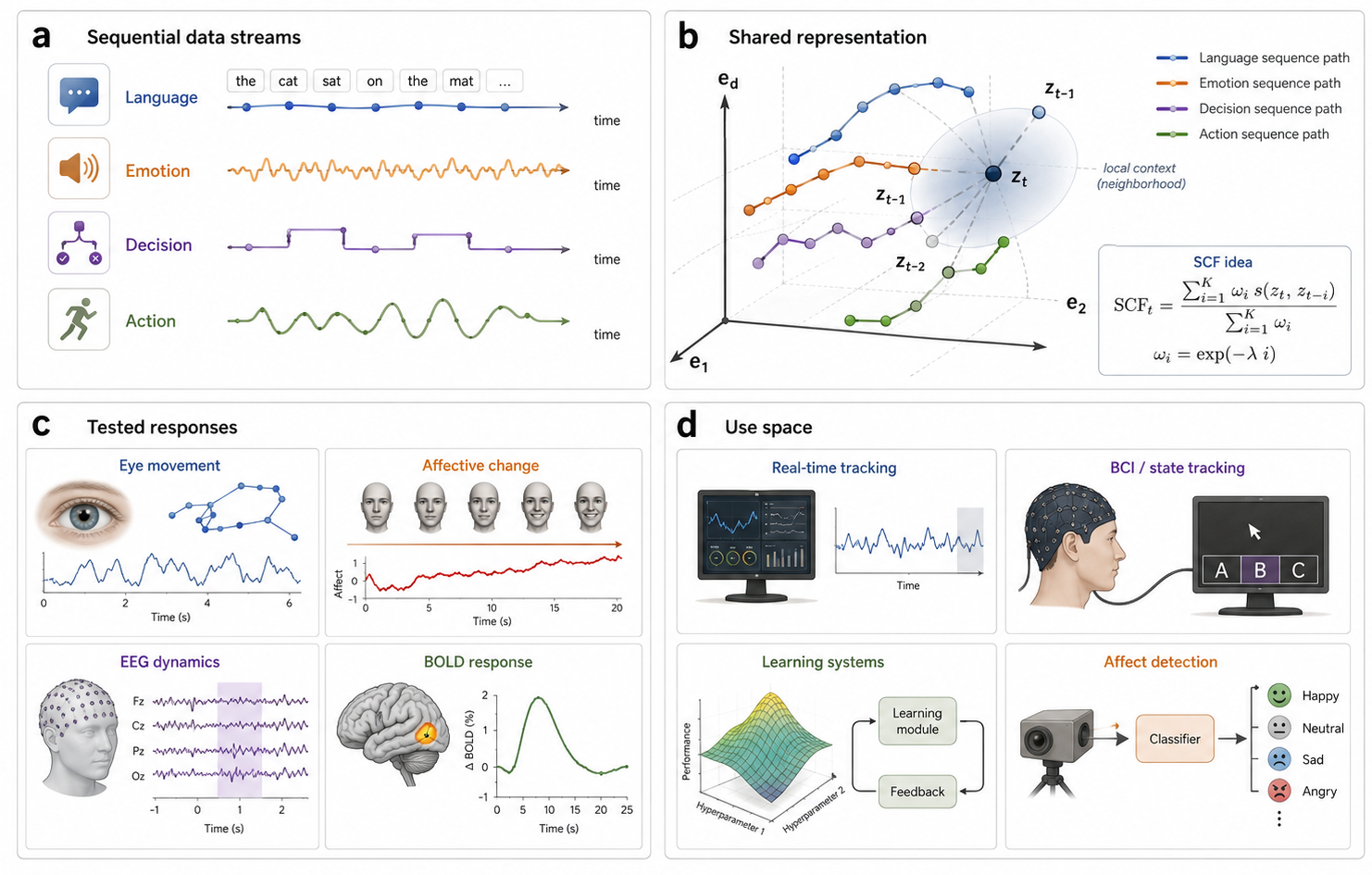}
	%  \caption{\textbf{Sequential contextual fit as a general metric.} (a) Ordered sequence units are mapped by a representation function into task-specific vectors. The resulting scalar SCF value is used as a model-ready predictor: low SCF indicates contextual mismatch or transition pressure, whereas high SCF indicates continuity or fluent integration. (b) In embedding space, the current state is compared with recent context states using similarity links weighted by a linear recency kernel. The 3D geometry is a visualization of a potentially high-dimensional domain-specific embedding space. (c) The weighted similarities are reduced to one interpretable score and entered into statistical models together with controls and random effects.}
	\caption{ SCF across sequential systems. Panel(a), Representative sequential data streams from language, emotion, decision and action domains. Panel(b), Domain-specific observations are mapped into domain-specific embedding spaces, where sequential contextual fit (SCF) quantifies the recency-weighted similarity between the current state and its preceding local context. Panel(c), SCF is evaluated against selected behavioural and neural responses, including eye movements, affective change, EEG dynamics and BOLD responses. Panel(d), Potential application areas include real-time monitoring, brain--computer interface and state tracking, adaptive learning systems and affect detection. SCF is intended as a general measurement layer whose empirical interpretation depends on the representation, sequence definition and validation design.}
	\label{fig:scf-framework}
\end{figure}

Figure~\ref{fig:scf-framework} summarizes how SCF maps diverse sequential data into domain-specific embedding spaces, yields a recency-weighted predictor of contextual fit, and relates this predictor to behavioral, neural and applied outcomes.

\section{Results}

%\subsection*{A common operator links local context to state transitions}
\subsection{SCF quantifies local contextual fit}

Across the analyses, we represented each ordered unit in a domain-appropriate vector space and explored whether the current state matched its recent context. Given a current unit $z_t$ and $K$ preceding units, SCF was defined as
\begin{equation}
\mathrm{SCF}_t = \frac{\sum_{i=1}^{K} w_i \cos(z_t, z_{t-i})}{\sum_{i=1}^{K} w_i},
\end{equation}
where $w_i$ is a linear recency kernel that assigns larger weights to more recent context units. Higher values indicate local representational continuity; lower values indicate contextual mismatch or transition pressure. The full metric definition, alternative operators and computational properties are described in Methods section. %#%; the primary Results use the direct recency-weighted specification.

Panel~(b) in Figure~\ref{fig:scf-framework} illustrates the core computation. The primary analyses use a multi-unit recent context, whereas the one-step variant serves as a diagnostic of whether broader context contributes beyond adjacent change.
%Figure~\ref{fig:scf-framework} summarizes the SCF computation, from ordered sequence units to embedding-space similarity and the resulting model-ready predictor. The primary analyses use a multi-unit recent context; the one-step case is evaluated separately as a diagnostic of whether multi-step context adds information beyond adjacent change.

%\subsection*{Semantic fit predicts reading, EEG and BOLD responses beyond surprisal}
\subsection{SCF beyond surprisal in language}

We first tested whether SCF explains language processing beyond probabilistic predictability. Language provided the largest testbed for this comparison, with complementary eye-movement, EEG and fMRI analyses \citep{sun2024attention1,Sun2025sentence,sun2026attention, sun2026speech,sun2026contextualsemanticrelevanceword,sun2026contextualsemanticrelevancetracks}.  In the Multilingual Eye-Movement Corpus, SCF predicted fixation durations across 13 languages while controlling surprisal, word frequency and word length. Higher SCF predicted shorter first fixation, gaze and total fixation durations, whereas higher surprisal generally predicted longer reading times (the threshold \textit{p} < 0.05, and it was applied in all GAMMs tests in the present study). The SCF effect strengthened for later measures: mean $F$ values increased from 10.35 for first fixation duration to 26.83 for gaze duration and 26.51 for total fixation duration in BERT-paired models, and from 10.37 to 22.37 and 23.16 in GPT-paired models.

\begin{figure}[!h]
    \centering
    \includegraphics[width=\textwidth]{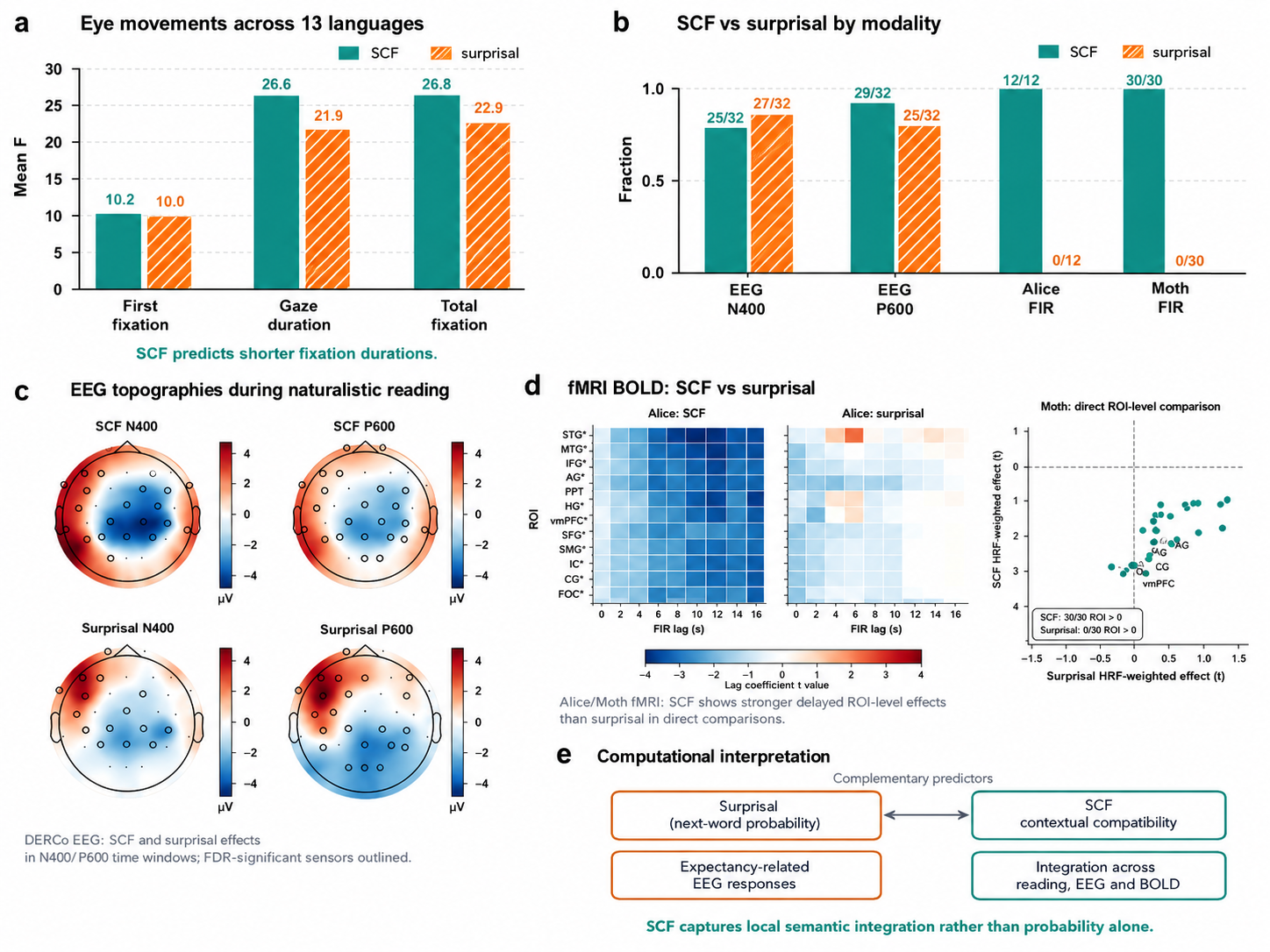}
    \caption{\textbf{SCF predicts language processing across behaviour and brain activity.} (a) Across the 13-language MECO eye-tracking analysis, SCF predicts first fixation duration, gaze duration and total fixation duration. (b) Summary of significant effects across EEG and fMRI analyses shows broad semantic-fit effects alongside the domain baseline, surprisal. (c) DERCo EEG topographic maps show semantic-relevance and surprisal effects over N400 and P600 windows, with FDR-significant sensors outlined. (d) Alice/Moth fMRI analyses show delayed FIR/HRF-weighted semantic-relevance effects across language-related ROIs, with weaker or absent matched surprisal effects in the direct comparison. (e) The language evidence supports SCF as a contextual-integration predictor complementary to next-word probability.}
    \label{fig:language-evidence}
\end{figure}

The EEG analysis showed that SCF also predicted word-locked neural responses during naturalistic reading. SCF showed broader FDR-corrected scalp coverage than surprisal in both the N400 and P600 windows, with effects in 23 of 32 and 25 of 32 channels, respectively, compared with 15 and 22 channels for surprisal. Channel-wise GAMMs and relative model comparisons gave convergent evidence, with the full distributions shown in Fig.~\ref{fig:language-evidence}.

The fMRI analyses extended this pattern to delayed hemodynamic responses during naturalistic speech comprehension. In the Alice dataset (Alice), FIR/deconvolution of original BOLD time series showed SCF effects in all 12 predefined ROIs after FDR correction, whereas surprisal was not significant in any ROI. Transformed-BOLD GAMMs showed semantic relevance effects in all 12 ROIs and surprisal effects in 6 of 12 ROIs. In the Moth dataset (Moth), HRF-weighted 4--12 s FIR/deconvolution showed consistent negative SCF effects in all 30 analyzable ROIs under both two-sided likelihood-ratio and directional FDR criteria, whereas surprisal did not show a comparable delayed pattern. In short, the language results show that SCF is a cross-measure predictor of language processing, complementary to surprisal rather than reducible to it (Fig.~\ref{fig:language-evidence}).

\subsection{SCF predicts affective and neural transitions}

The language results showed that SCF predicts processing in symbolic sequences beyond next-word probability. We next tested whether the same principle extends to continuous, non-symbolic affective signals. In the DEAM music-emotion dataset \citep{Aljanaki2017deam}, each song was segmented into 2 s units and represented using MERT audio embeddings \citep{li2024mert}. SCF was computed from the current unit and the preceding four units. The response was the absolute change in continuous affect at the next time point. The full GAMM included SCF, local acoustic change, current valence, current arousal, song random effects and rater random effects.

SCF robustly predicted next affective change, $F=89.13$, $p<2\times10^{-16}$, with $n=194{,}777$ observations. Current valence and arousal also contributed strongly, but the acoustic-change baseline did not explain comparable variance in the full model, $F=0.98$, $p=.248$. Within this same DEAM dataset, removing SCF increased AIC by 294.52, whereas removing acoustic change increased AIC by 0.31. These are conditional within-domain comparisons, not values on a common cross-domain effect-size scale. Lower SCF predicted larger subsequent affective change, indicating that emotion routes are sensitive to how well the current audio state fits the recent audio context.

We then asked whether stimulus-side SCF predicts neural-state updating during emotional audiovisual interaction. In the EAV dataset \cite{lee2024eav}, the full dataset contains 42 participants, each contributing 200 20 s interactions recorded with 30-channel EEG, audio and video. The present analysis used participants 1--10, comprising 2,000 interactions. After excluding the first four 1 s units in each interaction required to compute the context, 32,000 analyzable visual/EEG units remained. Each interaction was segmented into 1 s visual units. The midpoint frame of each unit was embedded with CLIP \cite{radford2021clip}, and stimulus-SCF was computed from the current visual embedding and the preceding four visual units. The response was the L2 transition of EEG features between the current 1 s EEG unit and the recency-weighted previous-four-unit EEG context. The full GAMM included SCF, one-step visual embedding change, previous EEG transition, time within interaction, task, emotion, subject random effects and interaction random effects. Because this is a ten-participant subset analysis, the EAV result is treated as a cross-stream proof of concept rather than a full-sample estimate.

\begin{figure}[p]
    \centering
    \begin{minipage}[t]{0.49\textwidth}
        \centering
        \includegraphics[width=\linewidth]{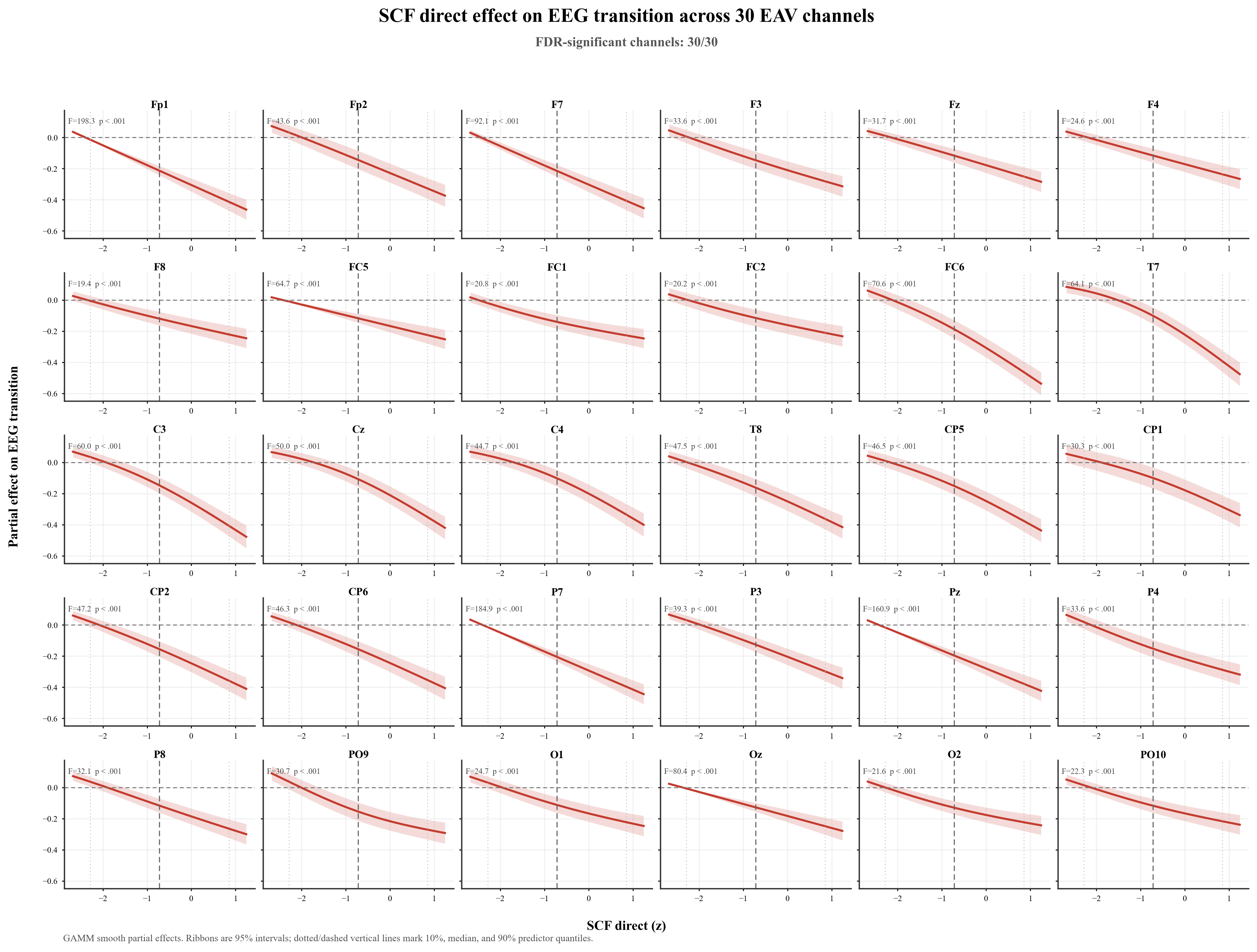}
    \end{minipage}\hfill
    \begin{minipage}[t]{0.49\textwidth}
        \centering
        \includegraphics[width=\linewidth]{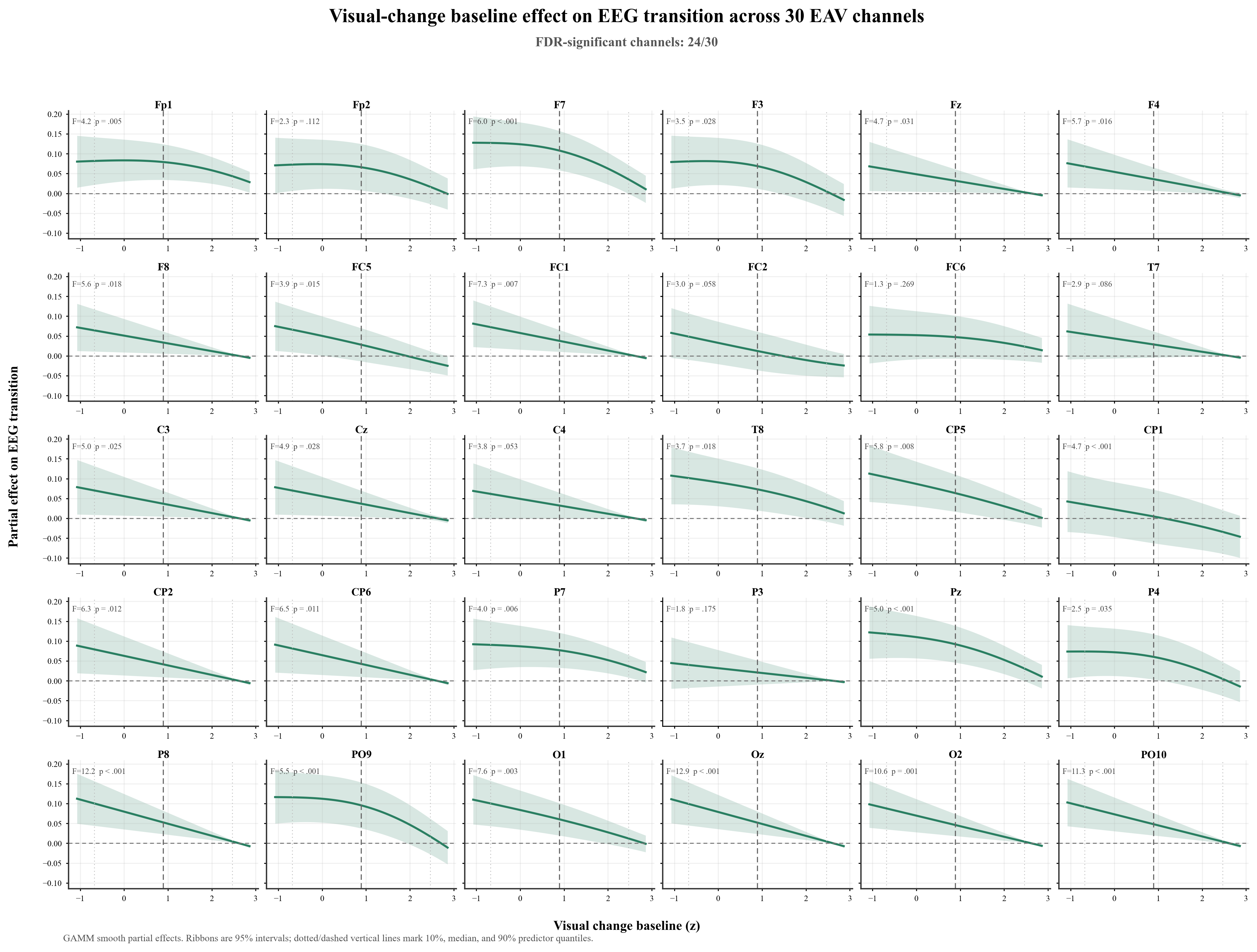}
    \end{minipage}

    \vspace{0.65em}
    \begin{minipage}[t]{0.49\textwidth}
        \centering
        \includegraphics[width=\linewidth]{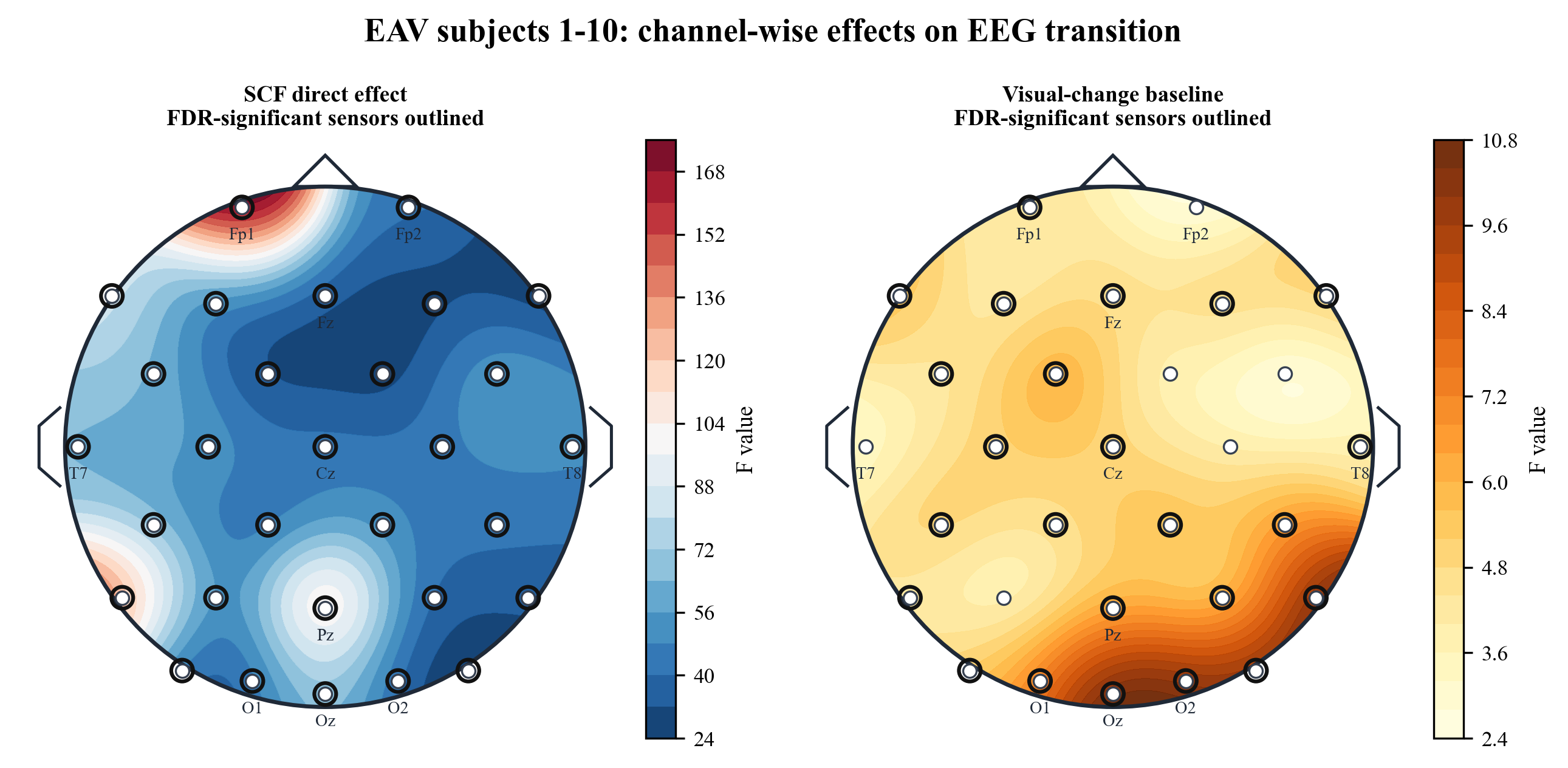}
    \end{minipage}\hfill
    \begin{minipage}[t]{0.49\textwidth}
        \centering
        \includegraphics[width=\linewidth]{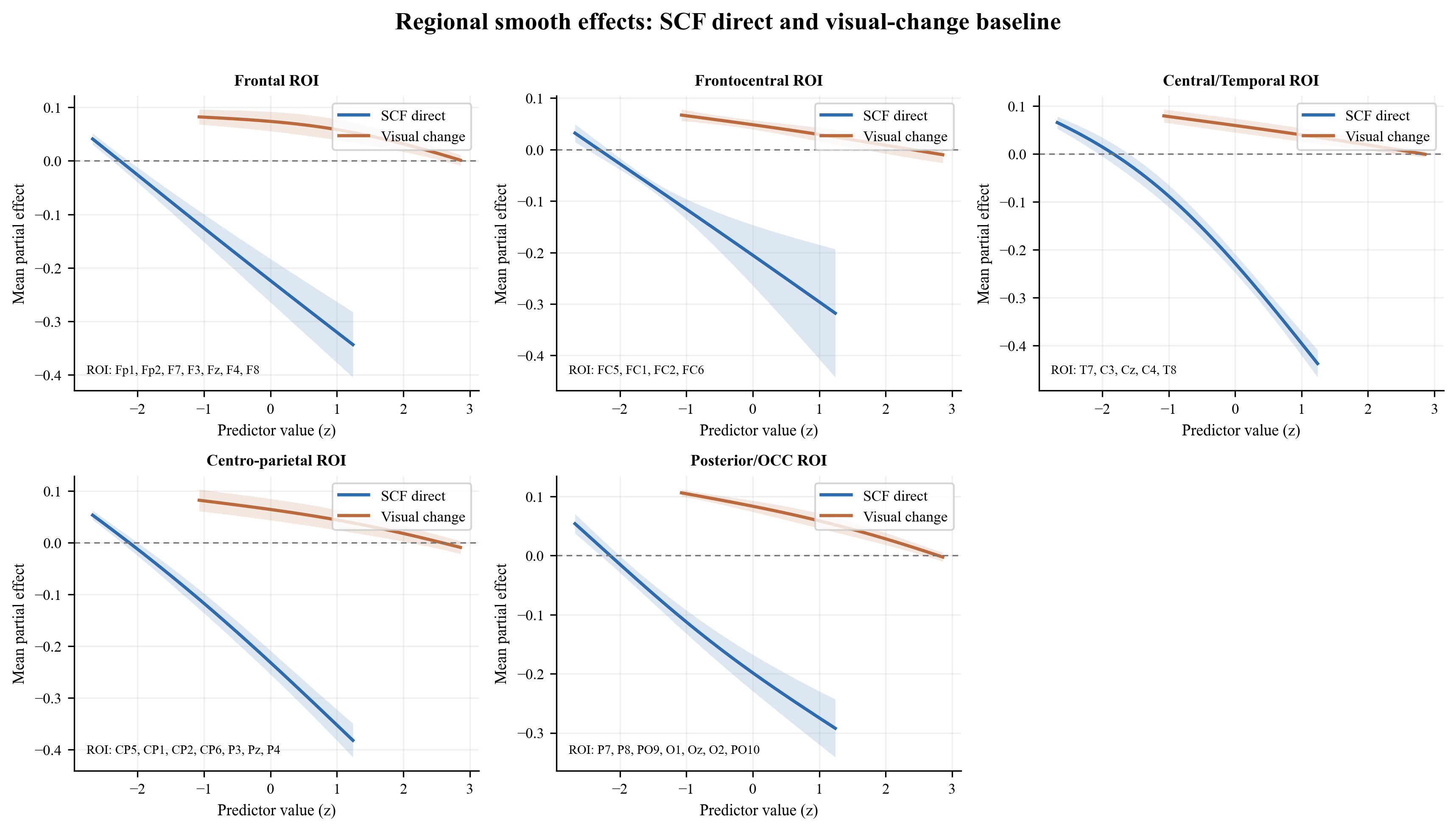}
    \end{minipage}
    \caption{\textbf{Visual stimulus-SCF predicts EEG-state transitions during audiovisual emotion processing.} EAV analyses used 1 s video/EEG units from participants 1--10 of the 42 available participants, yielding 32,000 analyzable units after context-window exclusion. The main predictor was CLIP-derived visual stimulus-SCF over the preceding four visual units; the baseline was one-step visual embedding change. (a,b) Channel-wise GAMM smooth partial effects for SCF direct and for one-step visual change. (c) Scalp-distribution summaries of channel-wise F statistics, with FDR-significant sensors outlined. (d) Regional smooth-effect summaries across frontal, frontocentral, central/temporal, centro-parietal and posterior/occipital channel groups. Channel-wise plots use tractable visualization models with subject random effects; the primary inferential model includes subject and interaction random effects.}
    \label{fig:eav-emotion-eeg}
\end{figure}

Across 32,000 units and 2,000 interactions, visual stimulus-SCF predicted EEG transition beyond the visual-change baseline and temporal controls. Within the EAV metric comparison, the unweighted direct SCF variant had the strongest conditional support, $F=91.30$, $p<2\times10^{-16}$, with $\Delta\mathrm{AIC}=234.42$ when removed from the same full model. The weighted direct SCF and pairwise-context variants were similarly supported, $F=76.80$ and $F=95.79$, respectively, both $p<2\times10^{-16}$. Visual change was also significant, $F=66.89$, $p<2\times10^{-16}$, but SCF remained predictive after this baseline was included. Channel-wise visualization models showed that SCF effects were spatially widespread, and 30/30 EAV channels survived FDR correction for the SCF effect, whereas 24/30 survived for visual change. These channel-wise models are used as scalp summaries rather than source localization, and the primary inference is the full mixed model with interaction random effects (Fig.~\ref{fig:eav-emotion-eeg}).

\subsection*{SCF predicts choice switching and activity transitions}

To test whether the same contextual-fit principle extends to non-linguistic motor sequences, we analysed the UCI Human Activity Recognition Using Smartphones (HAPT) dataset \citep{reyesortiz2016transition}. The HAPT dataset contains accelerometer and gyroscope recordings segmented into 2.56 s windows and annotated with both ongoing activities and postural transitions. It therefore provides a direct test of whether the relation between a current state and recent context predicts an upcoming change in action state, rather than only variation in sensor amplitude. Each window was represented by 561 engineered sensor features and a 32-dimensional bottleneck embedding from an unsupervised autoencoder. SCF compared the current embedding with the preceding five windows from the same subject/split sequence. The binary response indicated whether the labelled state involved a postural transition. The full binomial GAMM included direct sensor change, previous activity, window index, split and a subject random effect. The current activity label was excluded because it directly defined the transition response.% Note that the current activity label was excluded because it directly defined the transition response.

\begin{figure}[!h]
    \centering
    \includegraphics[width=\textwidth]{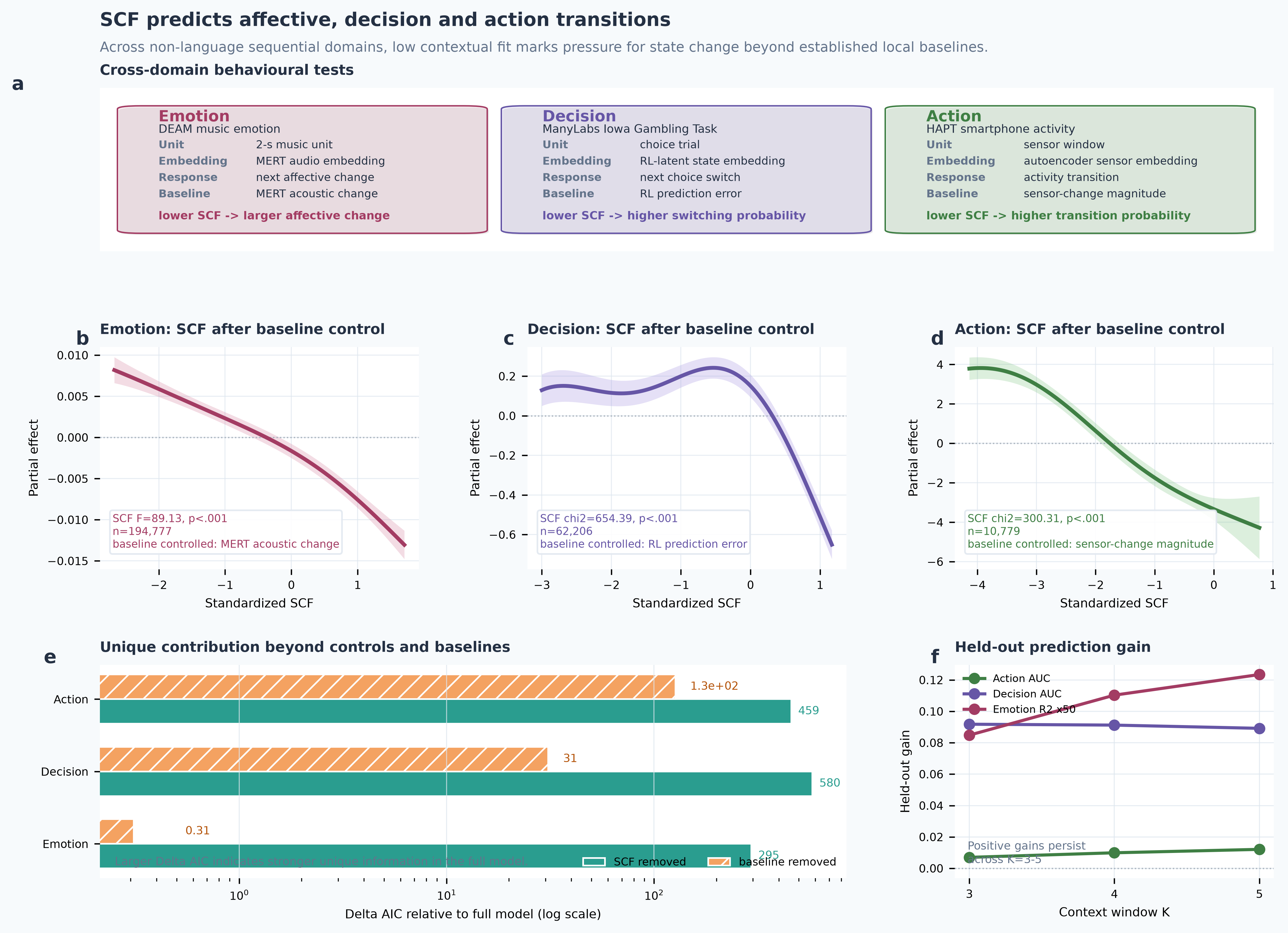}
    \caption{\textbf{SCF predicts affective, decision and action transitions.} (a) Cross-domain design for the non-language behavioural analyses, showing the dataset, sequence unit, embedding model, response variable and domain baseline for emotion, decision making and action. (b--d) GAMM partial-effect curves show the SCF effect after controlling for the corresponding domain baseline: MERT acoustic change for DEAM emotion, RL prediction error for Iowa Gambling Task decisions and sensor-change magnitude for HAPT action. (e) Full-versus-reduced within-domain comparisons show the conditional contribution of SCF beyond controls and baselines. (f) Leave-subject-out prediction gains persist across recent-context window sizes $K=3$--5.}
    \label{fig:behavioural-transitions}
\end{figure}

SCF significantly predicted postural transitions in the full binomial GAMM, $\chi^2=300.31$, $p<.001$, while the model explained 76.5\% of the deviance. The direct SCF effect was significant but non-monotonic, whereas the centroid-$L_2$ variant provided the clearest monotonic association with activity transitions. %Lower SCF was associated with a higher probability of an activity-state transition. 
 Direct sensor change also remained significant, $\chi^2=88.55$, $p<.001$, indicating that the SCF effect was not reducible to low-level signal variation. Removing SCF increased AIC by 458.79, compared with 126.78 after removing sensor change, supporting a substantial conditional contribution of SCF within the HAPT analysis.

\subsection{Decision-state fit predicts neural-state updating}

In the Iowa Gambling Task analysis \citep{chavez2026iowa}, each trial was represented by a reinforcement-learning latent state vector, and the response was whether the participant switched choices on the next trial. The baseline was negative reinforcement-learning prediction error. The full binomial GAMM included SCF, negative prediction error, current net outcome, trial number, current choice, participant random effects and study random effects. SCF strongly predicted next-trial switching, $\chi^2=654.39$, $p<2\times10^{-16}$, in $n=62{,}206$ trials. Prediction error also contributed, $\chi^2=36.98$, $p=4.64\times10^{-6}$, but, within this same IGT model, SCF had the larger conditional model-fit contribution: removing SCF increased AIC by 580.25, whereas removing prediction error increased AIC by 30.91. This comparison is conditional on the IGT controls and is not a cross-domain effect-size comparison.

Having established that contextual fit predicts affective, choice and activity transitions, we next investigated whether the same relation also predicts neural updating during decision making. We analysed an independent decision-making EEG dataset. Behavioural SCF was computed from the decision-state sequence using the behavioural/RL-latent representation, whereas the neural response was computed independently from the baseline-corrected 0--2 s EEG segment following the task marker. Single-trial EEG segments were downsampled and embedded using PCA; neural transition magnitude was defined as the L2 distance between the current EEG embedding and the recency-weighted EEG context over the preceding five trials. The full GAMM included behavioural/RL-latent SCF, negative prediction error, current net outcome, loss indicator, trial number, choice and participant random effects.

SCF predicted EEG transition magnitude, $F=5.83$, $p=.00196$, and removing SCF increased AIC by 7.38. Negative prediction error did not significantly predict this neural response, $p=.244$. Lower decision-state SCF was associated with larger 0--2 s EEG transitions, consistent with greater neural updating when the current decision state poorly fitted the recent decision context.

\begin{figure}[!h]
    \centering
    \includegraphics[width=\textwidth]{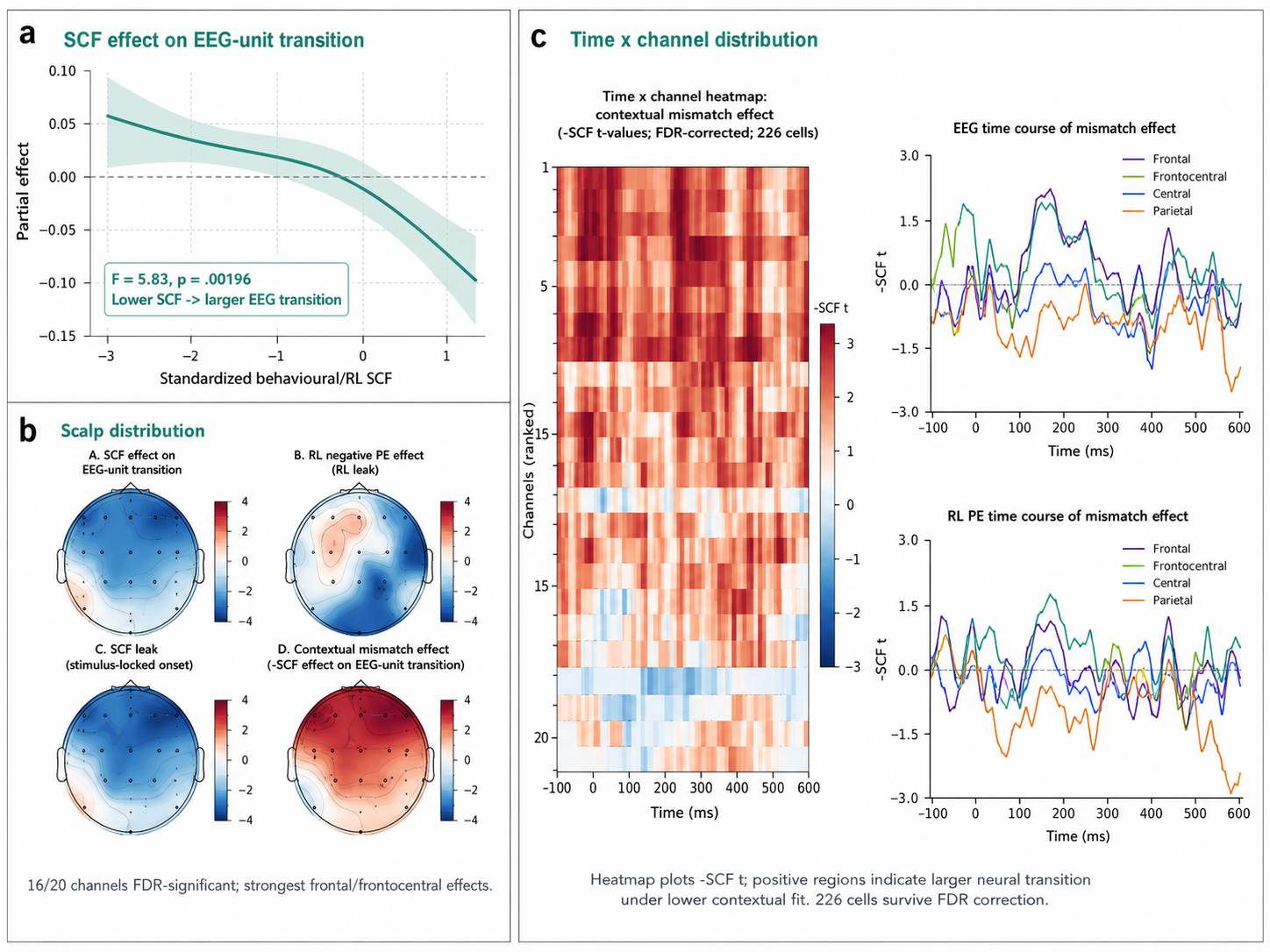}
    \caption{\textbf{Decision-related SCF predicts distributed EEG-state updating.}
\textbf{a}, Estimated partial effect of standardized behavioural/RL SCF on EEG-unit transition. Lower SCF was associated with larger neural-state transitions ($F=5.83$, $p=.00196$); the shaded region denotes the confidence interval and the dashed horizontal line marks zero effect.
\textbf{b}, Scalp distributions for the SCF effect on EEG-unit transition, the negative reinforcement-learning prediction-error effect, stimulus-locked SCF leakage and the contextual-mismatch effect, defined as the negative SCF effect on EEG-unit transition. Sixteen of 20 channels survived FDR correction, with the strongest effects over frontal and frontocentral regions.
\textbf{c}, Time-by-channel distribution of the contextual-mismatch effect. The heatmap shows FDR-corrected negative-SCF $t$ values across ranked EEG channels and time, with positive values indicating larger neural transitions under lower contextual fit; 226 channel--time cells survived FDR correction. The line plots show the corresponding EEG and reinforcement-learning prediction-error time courses for frontal, frontocentral, central and parietal channel groups.}
\label{fig:decision-eeg-transition}
   % \caption{\textbf{Contextual mismatch predicts neural-state updating during decision making.} (a) Route B treats each Iowa Gambling Task trial as one 0--2 s EEG unit and computes neural-state updating as the L2 transition between the current EEG embedding and the recency-weighted previous-five-trial EEG context. (b) Behavioural/RL-latent SCF predicts EEG-unit transition in the full GAMM after controlling RL prediction error, outcome, loss, trial, choice and participant effects. (c) Model evidence shows a stronger smooth effect for SCF than for RL prediction error in the same fitted model, with reduced-model support when SCF is removed. (d,e) A coordinated neural-evidence band shows the scalp distribution and time $\times$ channel distribution of the effect: 16/20 channels and 226 time $\times$ channel cells survive FDR correction.}
   
    \label{fig:decision-eeg}
\end{figure}

Channel-wise analyses showed broadly distributed SCF effects, with the strongest effects over frontal and frontocentral electrodes. Sixteen of 20 channels survived FDR correction at $q<.05$, including F4, $\beta=-0.0406$, $t=-4.93$, FDR $p=1.62\times10^{-5}$; F8, $\beta=-0.0394$, $t=-4.80$, FDR $p=1.62\times10^{-5}$; and F7, $\beta=-0.0367$, $t=-4.41$, FDR $p=6.89\times10^{-5}$. A time-by-channel analysis identified 226 FDR-significant SCF cells across 1,280 tests, with the strongest effects over frontal channels around 469 ms, 969 ms, 1063 ms, 1438--1469 ms and 1938 ms. These analyses are not source-localization results, but they indicate that contextual mismatch predicts distributed neural updating beyond prediction error and outcome-related controls (Fig.~\ref{fig:decision-eeg}).

%\subsection*{SCF generalizes across windows and participants}
\subsection{SCF is robust across context windows and held-out individuals}

%To separate the contribution of the embedding representation from that of the SCF operator, we compared the primary recency-weighted direct SCF with unweighted, pairwise-context and centroid-based variants computed from the same embeddings. Their predictive strength differed across domains, indicating that embedding availability alone was insufficient and that temporal ordering, recency weighting and current-to-context geometry were relevant to predictive performance.%Their predictive strength differed across domains, indicating that embedding availability alone was insufficient and the temporal ordering, recency weighting and current-to-context geometry also contributed to the observed effects.

To disentangle the contribution of the embedding representation from that of the SCF operator, we compared the primary recency-weighted direct SCF measure with unweighted, pairwise-context and centroid-based variants derived from the same embeddings. We also evaluated multiple embedding types. Predictive performance varied across domains, embedding representations and operator variants, indicating that access to embeddings alone was insufficient. Instead, temporal ordering, recency weighting and the geometry of the current state relative to its recent context contributed meaningfully to predictive performance. More details are presented in the Supplementary Material.

We first compared the one-step case ($K=1$) with the multi-unit windows used in the main analyses. In the action model, SCF remained significant at $K=1$, $\chi^2=241.74$, $p<.001$, and improved held-out AUC from 0.9743 to 0.9784; the corresponding improvement at $K=5$ was larger, from 0.9743 to 0.9864. In decision making, the one-step model also improved held-out AUC from 0.6482 to 0.7207, whereas the $K=3$--5 models improved it to 0.7359--0.7391. In emotion, the one-step smooth was not supported, $F=1.17$, $p=.380$, and the held-out $R^2$ gain was negligible (0.00002). The multi-unit models were supported for all tested windows, with $F=69.00$--$77.64$ and held-out $R^2$ gains of 0.0017--0.0025. This pattern indicates that multi-step context is especially important for the affective trajectory analysis, while one-step compatibility remains informative for action and choice switching.

We next tested whether SCF effects depended on the particular choice of multi-unit context window by refitting the main emotion, decision and action models with $K=3$, $K=4$ and $K=5$ preceding units.
%We then tested whether SCF effects depended on an arbitrary multi-unit context-window size by refitting the main emotion, decision and action models with $K=3$, $K=4$ and $K=5$ preceding units.
 SCF was significant for every tested multi-unit window in all three domains. For action, the SCF statistic ranged from $\chi^2=171.14$ to $300.95$, all $p<.001$. For emotion, it ranged from $F=69.00$ to $77.64$, all $p<.001$. For decision making, it ranged from $\chi^2=847.83$ to $1012.81$, all $p<.001$.

\begin{figure}[!h]
    \centering
    \includegraphics[width=\textwidth]{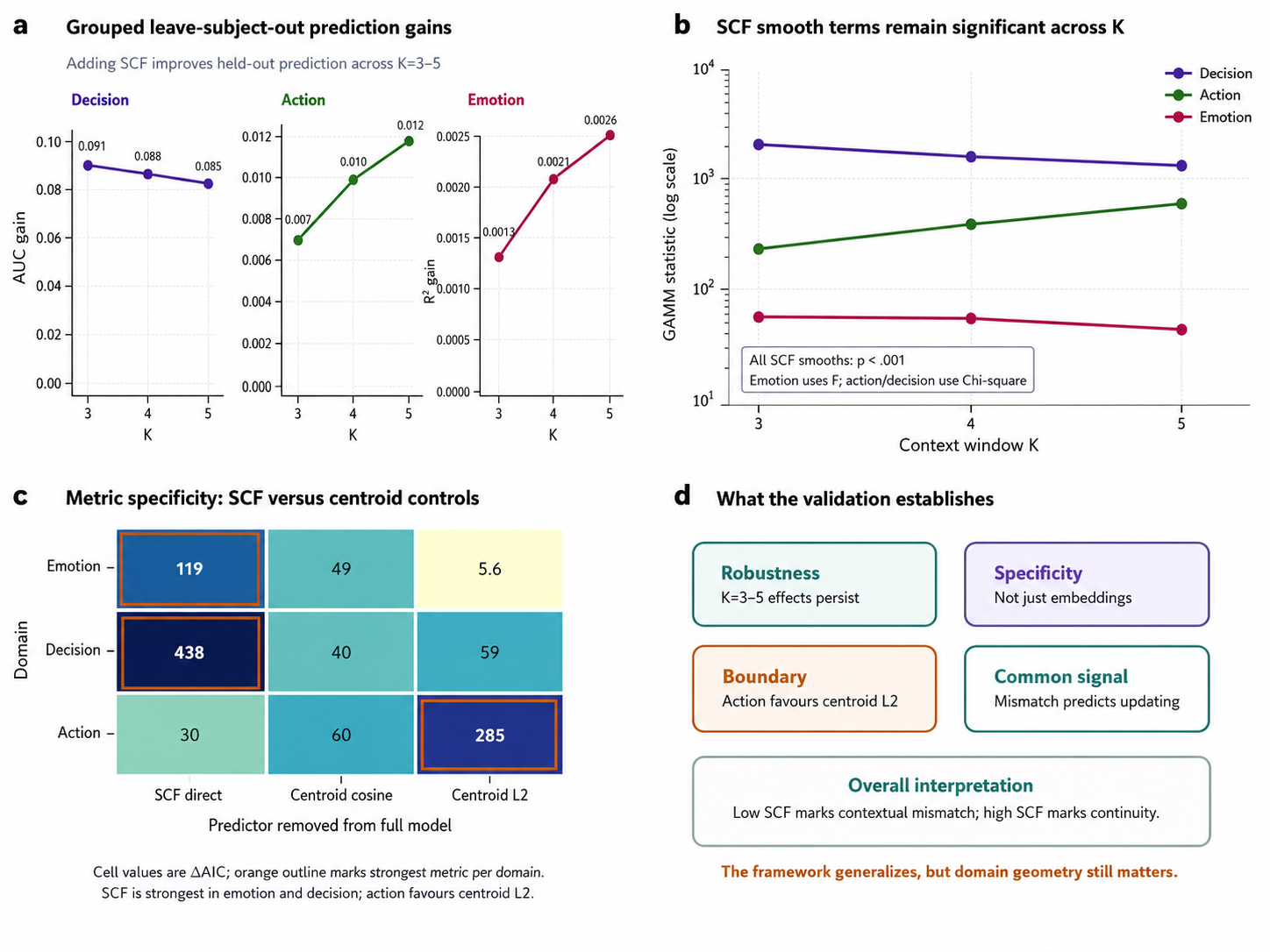}
   % \caption{\textbf{SCF effects are robust and metric-specific.} (a) Grouped leave-subject-out cross-validation shows that adding SCF improves held-out prediction across $K=3$--5 in decision making, action and emotion. (b) GAMM robustness tests show that SCF smooth terms remain significant across the same context-window sizes, with emotion reported as F statistics and action/decision as chi-square statistics. (c) Metric-specificity tests compare SCF direct with centroid cosine and centroid L2 metrics computed from the same embedding basis; reduced-model values are interpreted only within each dataset. The selected operator differs by domain, with centroid L2 providing an action-domain boundary condition. (d) The validation summary distinguishes robustness, metric specificity and domain boundary conditions while preserving the common interpretation that low SCF marks contextual mismatch or transition pressure.}
   \caption{\textbf{Robustness and metric specificity of sequential contextual fit.}
\textbf{a}, Grouped leave-subject-out prediction gains across context windows
$K=3$--$5$ for the decision, action and emotion datasets.
\textbf{b}, GAMM test statistics for SCF smooth terms across the same context
windows, showing that the effects remain statistically reliable despite changes
in $K$.
\textbf{c}, Metric-specific model comparisons based on $\Delta$AIC after
removing direct SCF, centroid-cosine or centroid-$L_2$ predictors from the full
model. Orange outlines indicate the strongest predictor within each domain.
\textbf{d}, Summary of the validation results, highlighting robustness across
context windows, specificity beyond generic embedding-centroid measures,
domain-dependent boundaries and a shared association between lower contextual
fit and greater state updating.}
    \label{fig:robustness-specificity}
\end{figure}

Grouped leave-subject-out cross-validation supported the same conclusion. Adding SCF improved held-out AUC in action from 0.9743 to 0.9813--0.9864 across $K=3$--5, and improved held-out AUC in decision making from 0.6468--0.6473 to 0.7359--0.7391. For emotion, baseline held-out $R^2$ values ranged from $-0.0022$ to $-0.0020$, and adding SCF produced $R^2$ values from $-0.0005$ to $0.0005$. 
In this sense, the results were not restricted to a single context-window specification, although predictive strength and the preferred operator remained domain-dependent (Fig.~\ref{fig:robustness-specificity}).
%Thus, SCF effects were not a single-window artifact, although the size and preferred operator remained domain-dependent (Fig.~\ref{fig:robustness-specificity}).

As the datasets contain many repeated observations, we interpreted significance together with model comparison and held-out prediction. Full-versus-reduced comparisons quantified the conditional contribution of SCF within each fitted dataset: removal increased AIC by $\Delta\mathrm{AIC}=294.52$ in DEAM emotion, $580.25$ in the gambling task, $458.79$ in HAPT action and $234.42$ for the best EAV embedding-context metric. These values are not compared across domains; they show incremental model evidence only relative to the controls, baseline and likelihood used in the same analysis.

\section{Discussion}

\subsection{Main findings}

Across the tested domains, SCF was associated with processing cost, behavioural or affective transitions and neural-state changes. Lower SCF generally corresponded to greater processing or updating, while the HAPT direct-SCF effect was non-monotonic and centroid L2 provided the clearest monotonic action-domain association. This pattern was observed in reading behaviour, language-related EEG and fMRI, music-evoked emotion, a subset of audiovisual emotion EEG, gambling decisions, activity transitions and decision-related EEG. The evidence is predictive rather than causal, but its convergence across different units, embedding spaces, response variables and datasets supports SCF as a useful candidate measurement of sequential compatibility. The language-EEG findings should not, however, be taken to identify SCF with either the N400 or P600 itself. Those analysis windows originate in distinct literatures on semantic incongruity and syntactic anomaly or reanalysis \citep{kutashillyard1980reading,osterhoutholcomb1992event}, whereas the present analyses test whether SCF predicts variation in neural responses within those windows.

%\subsubsection{Computational contributions}
SCF contributes a representation-level quantity that complements, rather than replaces, established predictors. Surprisal measures the negative log-probability of a linguistic unit given its preceding context \citep{hale2001probabilistic,levy2008expectation}; reinforcement-learning prediction error measures the discrepancy between expected and obtained outcomes \citep{schultz1997neural,garrison2013prediction}; and acoustic, visual or sensor change measures adjacent physical or representational discontinuity. SCF asks a different question: how well does the current state fit a recency-weighted local context? This operation is conceptually related to temporal-context models, which explains recency and contiguity through a dynamically evolving contextual state \citep{howardkahana2002distributed,polynnormankahana2009cmr}, and it shares with representational similarity analysis the use of distances or similarities to relate representations across computational, behavioural and neural measurements \citep{kriegeskorte2008representational}. It is also relevant to predictive-processing and event-segmentation accounts, in which violations of ongoing expectations can accompany perceptual or cognitive updating \citep{friston2010free,clark2013whatever,zacks2007eventperception}. However, SCF is an embedding-based computational operator, not a full generative theory, a literal neural prediction-error signal or a hidden-state sequence model. In particular, unlike a hidden Markov model, it does not estimate state-transition probabilities or infer a latent state sequence \citep{rabiner1989tutorial}.

The audiovisual emotion-EEG analysis provides especially important evidence for this distinction. We computed SCF from visual stimulus embeddings and constructed the response from EEG features, so the predictor and response were not algebraically derived from the same signal. Visual stimulus SCF remained informative after controlling for immediate visual change, previous EEG transition, time, task and emotion. This cross-modal separation reduces the risk of circularity, although it does not make the visual and neural streams statistically independent because both were recorded during the same interactions. 

%The action analysis also defines an important boundary condition. Direct SCF was predictive beyond sensor change, but centroid L2 provided the stronger embedding-context metric for activity transitions. This result suggests that action changes may be organized more by distance from a recent state prototype than by lag-specific current-to-context relatedness. This interpretation is post hoc and should be tested prospectively. Thus, the cross-domain claim is not that one operator must be optimal everywhere; it is that embedding-space compatibility and transition geometry can be tested with transparent, comparable operators.

The one-step sensitivity analysis further clarifies the role of context length. One-step SCF remained predictive for action and decision making, but multi-step windows gave larger held-out gains. In the emotion data, one-step SCF was not supported whereas the four-unit context was, suggesting that affective updating may depend more strongly on accumulated recent context than on adjacent acoustic similarity alone. This result does not establish a specific psychological integration window, but it is compatible with continuous-response research showing that perceived musical emotion evolves over time with changing musical features and with theoretical accounts in which musical emotion can arise through several interacting mechanisms, including expectancy \citep{schubert2004modeling,juslinvastfjall2008emotional}. It therefore supports the intended interpretation of SCF as a contextual-fit computational operator rather than a generic embedding-change score.

The decision-EEG findings should likewise be interpreted as prediction of trial-level neural-state change rather than identification of a canonical ERP component. Recent work reviewing IGT-ERP studies distinguishes evaluation, response selection and feedback processing as partially separable stages of decision making \citep{latibeaudiere2025decision}. The present 0--2 s transition measure aggregates activity over a broader interval and therefore cannot determine which specific ERP process carries the SCF association.

\subsection{Cognitive and neural implications}
Further, the theoretical significance of SCF is that it operationalizes a common question in sequential processing: can the current information state be integrated into the recently active representational context, or does it require an update of that state? SCF converts the relation between the current unit and its recent context into a deterministic compatibility score. It can therefore be computed directly as an interpretable signal of continuity or mismatch, whereas statistical models are needed to estimate how this signal relates to noisy behavioural, affective and neural responses, including nonlinearities, uncertainty and competing predictors. 

At the cognitive level, high contextual fit may support continuity and integration, whereas low fit may increase the need for attentional reallocation, representational reorganization, event segmentation or behavioural change. This interpretation is consistent with predictive-processing accounts in which mismatch drives updating, event-segmentation accounts in which transient prediction errors signal event boundaries, and temporal-context models in which recent context shapes continuity and contextual drift \citep{clark2013whatever,friston2010free,zacks2007event,howardkahana2002distributed}. The present findings therefore suggest that contextual compatibility may be an intermediate principle linking ongoing integration to cognitive and neural state transitions across domains. SCF does not demonstrate a single shared latent buffer or identify a specific neural circuit; rather, it provides evidence that the distinction between representational continuity and updating is reflected in processing time, affective change, choice switching, activity transitions and distributed neural-state variation. This representation-level interpretation is also compatible with similarity-based approaches that relate patterns of representation to behaviour and brain activity \citep{kriegeskorte2008representational}.

Several limitations qualify these findings. The analyses are secondary and support predictive associations, not causal claims. SCF may vary with the embedding, unit, context window, kernel and response definition, while the heterogeneous datasets preclude direct cross-domain comparison of effect magnitudes. Finally, $\Delta\mathrm{AIC}$ is conditional on each model's controls and is not a common effect-size scale. 

%\subsubsection{Outlook}

Beyond its role as an explanatory predictor, SCF offers a general measurement layer for systems that process ordered information. Once domain-specific units are embedded, the SCF calculation itself has cost \(O(Kd)\) per unit, which may support monitoring of contextual compatibility in applications such as brain--computer interfaces, adaptive learning, affective computing, human--machine interaction, recommendation systems and robotic activity monitoring. Low SCF can serve as an interpretable signal of contextual mismatch or likely state updating, whereas high SCF indicates continuity with recent context. Because SCF does not require a domain-specific probability, reward or latent-state model, it can also support cross-domain benchmarking and meta-analysis while remaining complementary to predictive and generative models.

In sum, the contribution of this study is not a new generative account of cognition or behaviour, but a candidate domain-general measurement operator. SCF makes current-to-recent contextual compatibility explicit, preserves recency and order, and produces a transparent predictor that can be tested alongside probability-based, value-based and physical-change baselines. The cross-domain evidence supports a common measurement framework. This combination of common structure and explicit boundary conditions is more informative than claiming that one metric is universally optimal.

\section{Conclusion}

Sequential contextual fit (SCF) provides a transparent, recency-weighted measure of how well a current information state matches its recent context. Across language, emotion, decision making, action and neural data, SCF showed predictive value beyond domain-specific baselines, while its effect shape and optimal representation remained domain-dependent. As a computational model, SCF offers a lightweight and interpretable measurement layer that can be applied across sequential systems without requiring a domain-specific probability, reward function or latent-state inference. Its low computational cost and domain-general formulation create opportunities for real-time state monitoring, brain--computer interfaces, adaptive learning and human--machine interaction. SCF is a useful tool for relating representational continuity and contextual mismatch to behavioural, affective and neural state variation.
%Future work should preregister the unit definition, embedding checkpoint, context window, kernel and baseline metrics before testing new datasets. It should also evaluate SCF in prospective held-out data, compare it with fitted temporal-context, hidden-state and event-boundary models, and test whether manipulating contextual compatibility changes subsequent processing or state updating. Because the SCF calculation is lightweight, it could eventually support real-time adaptive systems, but practical deployment will depend on the latency and reliability of embedding extraction and signal preprocessing. SCF may therefore serve as a measurement layer that complements, rather than replaces, generative models in computational neuroscience, behavioural science and multimodal sequential analysis.

\subsection*{Reproducibility}

For each domain, the project archive contains processed analysis datasets, small sample files for inspection, scripts for constructing embeddings and SCF predictors, GAMM scripts, reduced-model comparisons and plotting scripts. The main reproducible analysis chains are the HAPT autoencoder-SCF and sensor-change scripts for action, the DEAM MERT-SCF and acoustic-change scripts for music emotion, the EAV CLIP-SCF, visual-change and channel-wise emotion-EEG scripts, the RL-latent IGT scripts for decision making, the Route B trial-level EEG scripts for neural-state updating and the language SCF/surprisal scripts for reading, EEG and fMRI analyses. All reported results were generated from saved intermediate data and model-output files so that the figure panels and manuscript statistics can be traced back to the corresponding analysis records.

\subsection*{Software and computation}

Python workflows used Python 3.10.11 for data preparation, embedding extraction and figure generation. Statistical analyses used R 4.5.1 with the \texttt{mgcv} package (version 1.9.4) for GAMMs and standard R plotting and data-manipulation packages. The CLIP and MERT analyses used fixed pretrained checkpoints; the exact checkpoint names and analysis settings are recorded in the project scripts. Neural-network and embedding extraction steps used GPU acceleration where available, whereas GAMM fitting and reduced-model comparisons were run from saved intermediate analysis tables. The reported figures and statistics were generated from the saved analysis outputs rather than from manually edited values. Additional methodological details, domain-specific implementation choices, robustness analyses and metric-correlation diagnostics are provided in the \textbf{Supplementary Material (SM)}. The \textbf{SM} also reports supplementary figures, concurvity diagnostics, the effects of alternative SCF metrics, sensitivity analyses and data/code availability information supporting the results presented here.

%\section*{Competing interests}

%The author declares no competing interests.
% Precompiled bibliography for arXiv submission.

\end{document}